%% file: manuscript4update.tex
\documentclass[a4paper,fleqn]{cas-sc}
\usepackage[numbers]{natbib}
\usepackage{indentfirst} 
\def\tsc#1{\csdef{#1}{\textsc{\lowercase{#1}}\xspace}}
\tsc{WGM}
\tsc{QE}
\usepackage{amsmath,amssymb,amsfonts}%
\usepackage{amsthm}%
\usepackage{mathrsfs}%
\usepackage[title]{appendix}%
\usepackage{xcolor}%
\usepackage{textcomp}%
\usepackage{manyfoot}%
\usepackage{booktabs}%
\usepackage{algorithmicx}%
\usepackage{algpseudocode}%
\usepackage{listings}%
\usepackage[ruled,linesnumbered]{algorithm2e}
\usepackage{tikz}
\usetikzlibrary{shapes,arrows.meta,positioning}
\usepackage{graphicx}
\usepackage{subcaption}
\usepackage{caption}
\usepackage{siunitx}
\usepackage{tabularx} 
\usepackage{float}
\usepackage{comment}
\usepackage{tabto}
\usepackage[normalem]{ulem}
\usepackage{adjustbox}

\usepackage{textalpha}
\usepackage{placeins} 

\usepackage{tikz}

\usetikzlibrary{arrows.meta,positioning,fit,calc}

\tikzset{
  blk/.style={draw, rounded corners=2pt, align=center, font=\small,
              minimum height=10mm, minimum width=48mm, inner sep=3pt},
  blkS/.style={draw, rounded corners=2pt, align=center, font=\scriptsize,
               minimum height=8mm, minimum width=26mm, inner sep=2pt},
  layer/.style={draw, rounded corners=3pt, dashed, inner sep=6pt},
  arrow/.style={-Latex, thick},
  arrowThin/.style={-Latex, semithick},
  note/.style={font=\footnotesize, align=left},
}

\begin{document}
\let\WriteBookmarks\relax
\def\floatpagepagefraction{1}
\def\textpagefraction{.001} 

\shorttitle{Integrated Multi-Drone Task Allocation and Trajectory Generation in 3D Obstacle-Rich Environments}

\shortauthors{Yunes ALQUDSI, and Murat MAKARACI}

\title [mode = title]{Integrated Multi-Drone Task Allocation, Sequencing, and Optimal Trajectory Generation in Obstacle-Rich 3D Environments} 

\author[1,2]{Yunes ALQUDSI}[orcid=0000-0002-4246-9654]
\cormark[1] 
\ead{yunes.alqadasi@kocaeli.edu.tr}

\author[3]{Murat MAKARACI}

\affiliation[1]{organization={Aerospace Engineering Department, Faculty of Aeronautics and Astronautics, Kocaeli University},
            state={Kocaeli},
            country={Turkiye}}
\affiliation[2]{organization={Interdisciplinary Research Center for Aviation and Space Exploration, KFUPM},
            state={Dhahran},
            country={Saudi Arabia}}
            
\affiliation[3]{organization={Mechanical Engineering Department, Faculty of Aeronautics and Astronautics, Kocaeli University},
            state={Kocaeli},
            country={Turkiye}}
            
\cortext[1]{Corresponding author} 

\begin{abstract}
\abstractindent
Coordinating teams of aerial robots in cluttered three-dimensional (3D) environments requires a principled integration of \emph{discrete} mission planning---deciding which robot serves which goals and in what order---with \emph{continuous}-time trajectory synthesis that enforces collision avoidance and dynamic feasibility. This paper introduces IMD--TAPP (Integrated Multi-Drone Task Allocation and Path Planning), an end-to-end framework that jointly addresses multi-goal allocation, tour sequencing, and safe trajectory generation for quadrotor teams operating in obstacle-rich spaces. IMD--TAPP first discretizes the workspace into a 3D navigation graph and computes obstacle-aware robot-to-goal and goal-to-goal travel costs via graph-search-based pathfinding. These costs are then embedded within an Injected Particle Swarm Optimization (IPSO) scheme, guided by multiple linear assignment, to efficiently explore coupled assignment/ordering alternatives and to minimize mission makespan. Finally, the resulting waypoint tours are transformed into time-parameterized minimum-snap trajectories through a generation-and-optimization routine equipped with iterative validation of obstacle clearance and inter-robot separation, triggering re-planning when safety margins are violated. Extensive MATLAB simulations across cluttered 3D scenarios demonstrate that IMD--TAPP consistently produces dynamically feasible, collision-free trajectories while achieving competitive completion times. In a representative case study with two drones serving multiple goals, the proposed approach attains a minimum mission time of 136~s while maintaining the required safety constraints throughout execution.
\end{abstract}


\maketitle 


\section{Introduction}
Unmanned aerial vehicles (UAVs), including multirotor platforms, have evolved from laboratory prototypes into widely deployed autonomous systems for inspection, mapping, monitoring, and logistics. This progress has been enabled by advances in onboard sensing, computation, and control, together with the operational flexibility of vertical take-off and landing vehicles \cite{rojas2021unmanned,ref2}. Consequently, mission planning for UAVs is commonly studied through routing and path-planning formulations, and surveys summarize the breadth of algorithms proposed for UAV routing and for autonomous aerial operations in real environments \cite{alqudsi2025advancements, maity2023flying}.

Many practical deployments remain challenging for a single vehicle because of limited endurance, sensing range, and payload capacity. Coordinated multi-drone teams mitigate these limitations by distributing goals across robots, improving mission completion time, coverage, and robustness \cite{du2025survey}. At the same time, cooperation introduces tightly coupled challenges such as shared-airspace safety, congestion around obstacles, and the need for scalable coordination as the team size grows \cite{alqudsi2024coordinated}. Benchmarking and evaluation suites for swarm robotics highlight the importance of standardized, reproducible testing when comparing multi-robot coordination approaches \cite{ref4,ghanem2025decision}, and state-of-the-art reviews emphasize the remaining gaps between laboratory demonstrations and reliable, scalable field operation \cite{javed2024state}.

A persistent difficulty in multi-drone operation is guaranteeing safety while preserving efficiency. Collision avoidance must be enforced both with respect to static obstacles and between robots moving in the same airspace, often under dynamic constraints and imperfect information \cite{hafezi2022design}. Surveys of collision-avoidance schemes and autonomous drone swarms emphasize the importance of principled separation constraints and reliable replanning mechanisms that remain stable when multiple vehicles interact in close proximity \cite{rezaee2024comprehensive, saunders2024autonomous}. In parallel, research on distributed coordination and collective decision-making in robot swarms continues to develop mechanisms that support scalable group behavior under limited communication \cite{almansoori2024evolution,shorinwa2024distributed,wang2024decentralized}.

From a planning standpoint, multi-drone missions typically couple at least two interacting layers. At the discrete layer, the planner must assign goals to robots and determine feasible visit sequences (multi-robot task allocation) \cite{song2025comparative, dai2025heterogeneous}. At the continuous layer, the planner must generate collision-free trajectories that respect obstacle constraints, inter-robot separation, and vehicle dynamics \cite{alqudsi2024advanced}. While decoupling these layers can simplify implementation, it can also yield brittle solutions: an assignment that appears optimal under geometric distances may become infeasible once obstacles, timing, and dynamic limits are enforced. Related work in other domains has similarly argued for integrated schedule-and-trajectory optimization when conflicts arise in shared spaces \cite{yao2020integrated}. Metaheuristic optimization remains a practical choice for the resulting combinatorial search space, especially when the objective is a makespan and constraints must be checked repeatedly during evolution \cite{alqudsi2025injected, osaba2020soft,meng2007hybrid}.

This work addresses this coupling through the Integrated Multi-Drone Task Allocation and Trajectory Generation framework (IMD--TAPP), which builds environment-aware travel-cost matrices via graph search, optimizes assignments and visit orders via an injected particle swarm optimization (IPSO) algorithm guided by multiple linear assignment, and then synthesizes smooth, dynamically feasible trajectories with iterative safety validation and replanning. By combining environment-aware cost construction with discrete optimization and continuous-time trajectory synthesis, IMD--TAPP targets obstacle-rich 3D settings where both efficiency and safety are critical \cite{alqudsi2025integrated}.

\begin{figure}
\centering
\includegraphics[width=\textwidth]{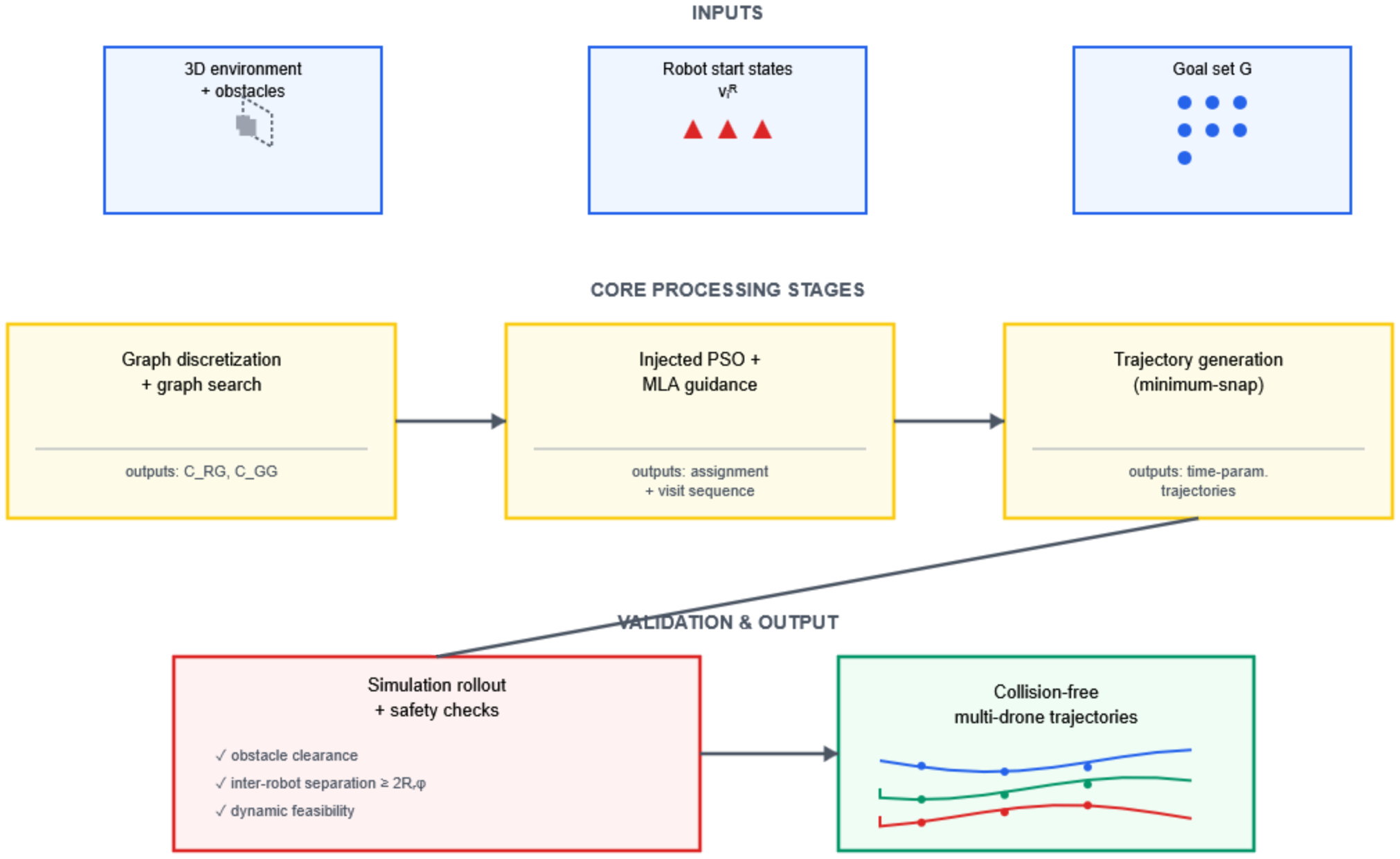}
\caption{The IMD--TAPP system schematic. The framework takes as inputs the 3D environment with obstacles, robot start states $v_R^i$, and goal set $G$. The core processing stages include: (i) graph discretization and search to compute cost matrices $\mathbf{C}_{RG}$ and $\mathbf{C}_{GG}$, (ii) IPSO with MLA guidance to optimize assignment and visit sequences, and (iii) minimum-snap trajectory generation to produce time-parameterized trajectories. The final stage validates obstacle clearance and inter-robot separation ($\geq 2R_r\phi$) through simulation rollout, outputting collision-free multi-drone trajectories.}
\label{fig:system-schematic}
\end{figure}

The main contributions of this paper are: (1) an obstacle-aware cost-construction module that uses 3D graph search to populate robot-to-goal and goal-to-goal travel-cost matrices in cluttered environments; (2) a joint assignment-and-sequencing optimizer based on IPSO with multiple linear assignment (MLA) guidance, improving convergence toward low-makespan solutions under the visit-once constraint; and (3) an end-to-end planning-to-trajectory workflow that generates optimal trajectories and iteratively revalidates obstacle clearance and inter-robot separation, triggering local replanning when safety constraints are violated.
A detailed system schematic for the work presented in research is provided in Figure~\ref{fig:system-schematic}, showing the inputs, core processing stages, and validation loop of the complete IMD--TAPP framework.

The remainder of the paper is organized as follows. Section~\ref{sec:related} reviews representative work on multi-robot task allocation and multi-agent path planning. Section~\ref{sec:problem} formalizes the IMD--TAPP problem, objective, and safety constraints. Section~\ref{sec:method} presents the proposed framework. Sections~\ref{sec:setup} and \ref{sec:results} describe the simulation setup and discuss the results. Section~\ref{sec:conclusion} concludes the paper and summarizes limitations and future research directions.

\section{Related Work}
\label{sec:related}
Multi-robot task allocation (MRTA) has been studied extensively, with foundational work providing taxonomies that distinguish problem variants by robot heterogeneity, task structure, and assignment coupling \cite{gerkey2004formal, ghauri2024review}. In the context of UAVs and autonomous flying robots, MRTA is frequently coupled to routing because feasibility and cost depend strongly on travel distances through obstacle fields and on timing constraints \cite{chen2024distributed}. Surveys of UAV routing and multi-UAV mission planning emphasize that coupling allocation decisions with environment-aware travel costs is essential for realistic performance evaluation \cite{rojas2021unmanned,song2023survey, peng2021review} and has motivated distributed MRTA strategies for scalability \cite{shorinwa2024distributed}.

A related line of research concerns the routing formulations that underlie multi-goal allocation problems. Multiple-traveling-salesman and vehicle-routing variants have been investigated through mathematical programming and heuristic approaches, including formulations and algorithms that address assignment, sequencing, and feasibility constraints \cite{ref20,ref16,ref8}. Dynamic variants incorporating additional limitations, such as energy constraints, further motivate algorithms that balance solution quality with computational tractability \cite{ref21, alqudsi2025towards}.

Optimization methods for combinatorial multi-robot planning span exact solvers, heuristics, and soft-computing techniques. Population-based methods are frequently adopted when exact formulations are impractical at scale or when repeated feasibility checks are required under complex constraints \cite{osaba2020soft}. Swarm-inspired coordination and hybrid metaheuristics have also been explored for distributed settings \cite{meng2007hybrid}, and recent work on collective decision-making highlights how adaptable mechanisms can support scalable group behavior in robot swarms \cite{almansoori2024evolution}.

Safe multi-agent navigation is equally critical in shared 3D environments. Many coordination problems can be mapped to multi-agent pathfinding (MAPF) on graphs, which plans conflict-free paths for multiple agents while optimizing criteria such as makespan \cite{stern2019multi}. For large teams, scalability and safety are central concerns, and prior work has investigated safe, scalable, and complete motion planning for large groups of interchangeable robots \cite{ref3}. In the UAV context, complete flight trajectory planning for multiple vehicles has also been studied, providing useful insights into feasibility under constraints \cite{burger2013complete}. Collision avoidance remains a persistent challenge in practice, and surveys review sensing, planning, and avoidance strategies as well as open issues \cite{hafezi2022design,rezaee2024comprehensive,saunders2024autonomous}.

Integrated task assignment and path planning has therefore attracted increasing attention. Distributed and decentralized strategies have been explored to support coordination under limited communication, dynamic conditions, and real-time replanning requirements \cite{wang2024decentralized,du2023multi,li2023dynamic}. Recent studies examine synergistic coupling of allocation with obstacle-aware planning \cite{abro2024synergistic} and learning-based coupling of assignment and navigation in dynamic obstacle environments \cite{kong2024multi}. More broadly, related research in shared conflict zones has demonstrated the value of integrating scheduling and trajectory optimization in a single decision loop \cite{yao2020integrated}, which is conceptually aligned with the motivation of IMD--TAPP.

Finally, trajectory generation for quadrotors commonly relies on polynomial optimization to obtain smooth and dynamically feasible flight profiles; minimum-snap formulations are particularly influential because they promote smoothness and enable constraint enforcement along waypoint corridors \cite{mellinger2011minimum}. Complementary work has developed numerically stable trajectory generation and general optimization frameworks for highly maneuverable multirotor drones in complex environments \cite{alqudsi2023numerically,alqudsi2023general}, and integrated task assignment with trajectory generation has been investigated to improve collision avoidance and flight efficiency in multi-drone operations \cite{alqudsi2024enhancing}. IMD--TAPP follows this broad philosophy by combining graph-search cost construction with discrete optimization and continuous-time trajectory synthesis, while maintaining an explicit safety-validation loop.

\section{Problem Formulation}\label{sec:problem}
\begin{figure}
\centering
\includegraphics[width=0.75\textwidth]{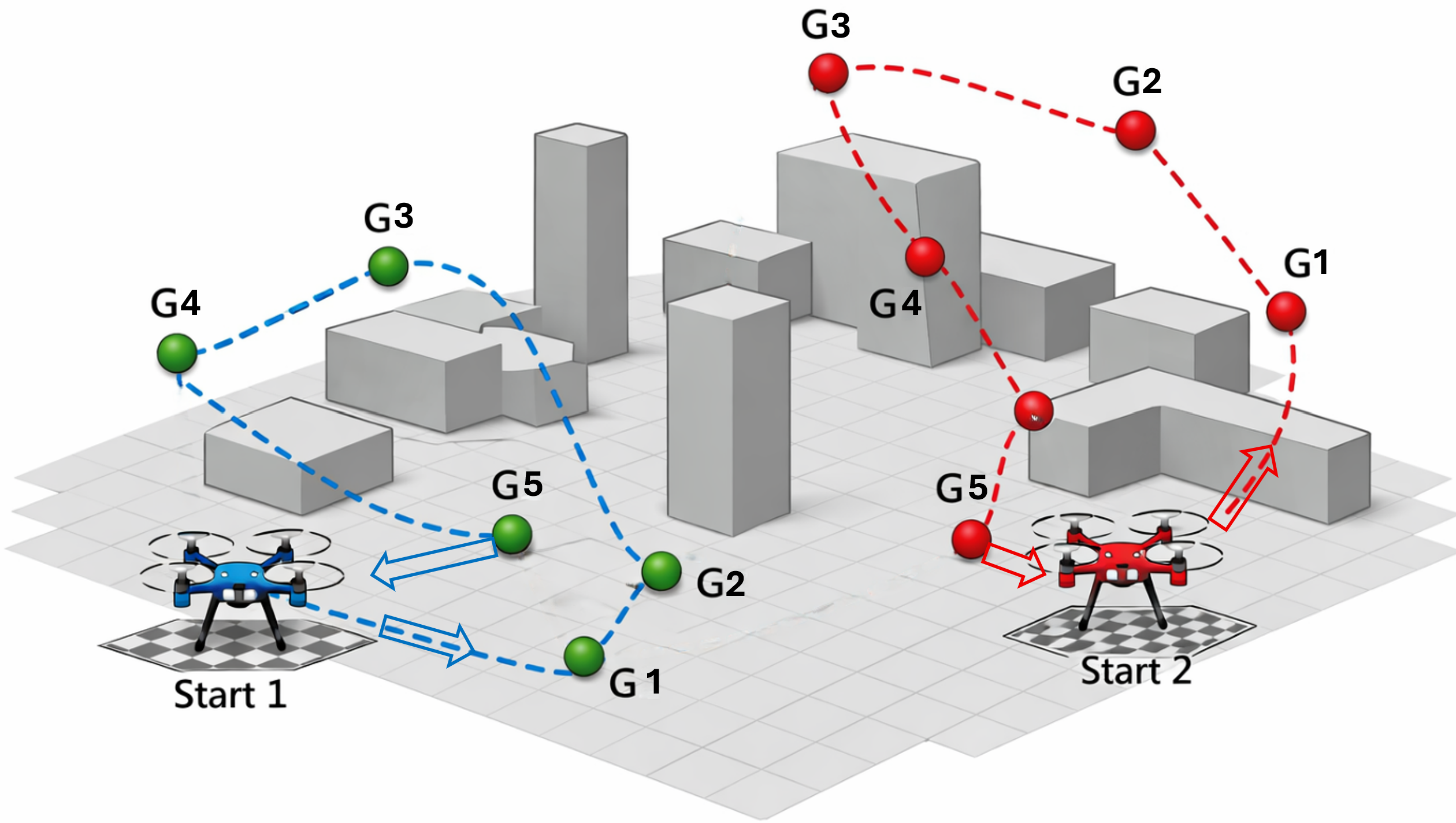}
\caption{Illustrative IMD--TAPP scenario: two drones must cooperatively visit seven goals in a 3D environment with obstacles, with each goal visited exactly once and each drone returning to its start location.}\label{fig01}
\end{figure}

We consider a team of $R_N$ aerial robots that must collectively visit a set of $G_N$ goals in a bounded 3D workspace containing obstacles. Each goal must be visited exactly once by any robot, and each robot must return to its takeoff point after completing its assigned goals (Figure~\ref{fig01}).

The robots are treated as interchangeable with respect to goal servicing, and the environment is modeled as obstacle-rich and potentially challenging for line-of-sight motion.

Let $Sq_i$ denote the goal sequence assigned to robot $i$.

\begin{equation}
    \label{eq:eq2}
    \psi_{ij} = 
    \begin{cases}
        1 & \text{if goal $j$ is assigned to drone $i$ } \\
        0 & \text{otherwise}
    \end{cases}
    \end{equation}
 The traversal cost of a robot tour is denoted by $C(v_R^i \rightarrow Sq_i \rightarrow v_R^i)$ and is computed from obstacle-aware geometric path costs between successive waypoints, including (i) travel from the robot start $v_R^i$ to the first goal, (ii) travel between consecutive goals in $Sq_i$, and (iii) return from the last goal back to $v_R^i$. The IMD--TAPP objective is to minimize the mission makespan, i.e., the maximum tour cost among all robots:
\begin{equation}
\label{eq:obj}
J^\ast = \min_{Sq_1,\ldots,Sq_{R_N}} \; \max_{i \in \{1,\ldots,R_N\}} \; C\!\left(v_R^i \rightarrow Sq_i \rightarrow v_R^i\right).
\end{equation}

In many multi-goal aerial missions, each drone is required to terminate at a specified home or recovery location, which influences both allocation and timing decisions.

where $v_R^i$ is the initial state of robot $i$ and $Sq_i$ denotes its ordered list of assigned goals (possibly empty).

For feasibility, robot motion must satisfy obstacle avoidance and inter-robot safety. Collision avoidance between robots $i$ and $j$ is enforced through the distance constraint
\begin{equation}
\label{eq:eq3}
\| \eta_i(t) - \eta_j(t) \| \geq 2 R_r \phi, \quad \phi > 1, \quad t \in [t_0, t_f],
\end{equation}
where $\eta_i(t)$ is the position of robot $i$ at time $t$, and $R_r$ is a spherical safety radius that conservatively approximates the robot body. In addition, dynamic feasibility must be considered when generating time-parameterized trajectories, including bounds on velocities and accelerations \cite{alqudsi2023general}.

To improve readability, Table~\ref{tab:notation} summarizes the main symbols used in the formulation and algorithm description.

\begin{table}[t]
\caption{Key notation used in the IMD--TAPP formulation.}
\label{tab:notation}
\centering
\begin{tabular}{ll}
\toprule
Symbol & Meaning \\
\midrule
$R_N$ & number of robots (drones) \\
$G_N$ & number of goals \\
$Sq_i$ & ordered goal sequence assigned to robot $i$ \\
$C(\cdot)$ & travel cost computed from geometric paths \\
$J^\ast$ & optimal mission objective (makespan) \\
$\eta_i(t)$ & position of robot $i$ over time \\
$R_r$ & conservative robot safety radius \\
$\phi$ & safety inflation factor ($\phi>1$) \\
\bottomrule
\end{tabular}
\end{table}

\section{Proposed IMD--TAPP Framework}
\label{sec:method}
\begin{figure}
\centering
\includegraphics[width=\textwidth]{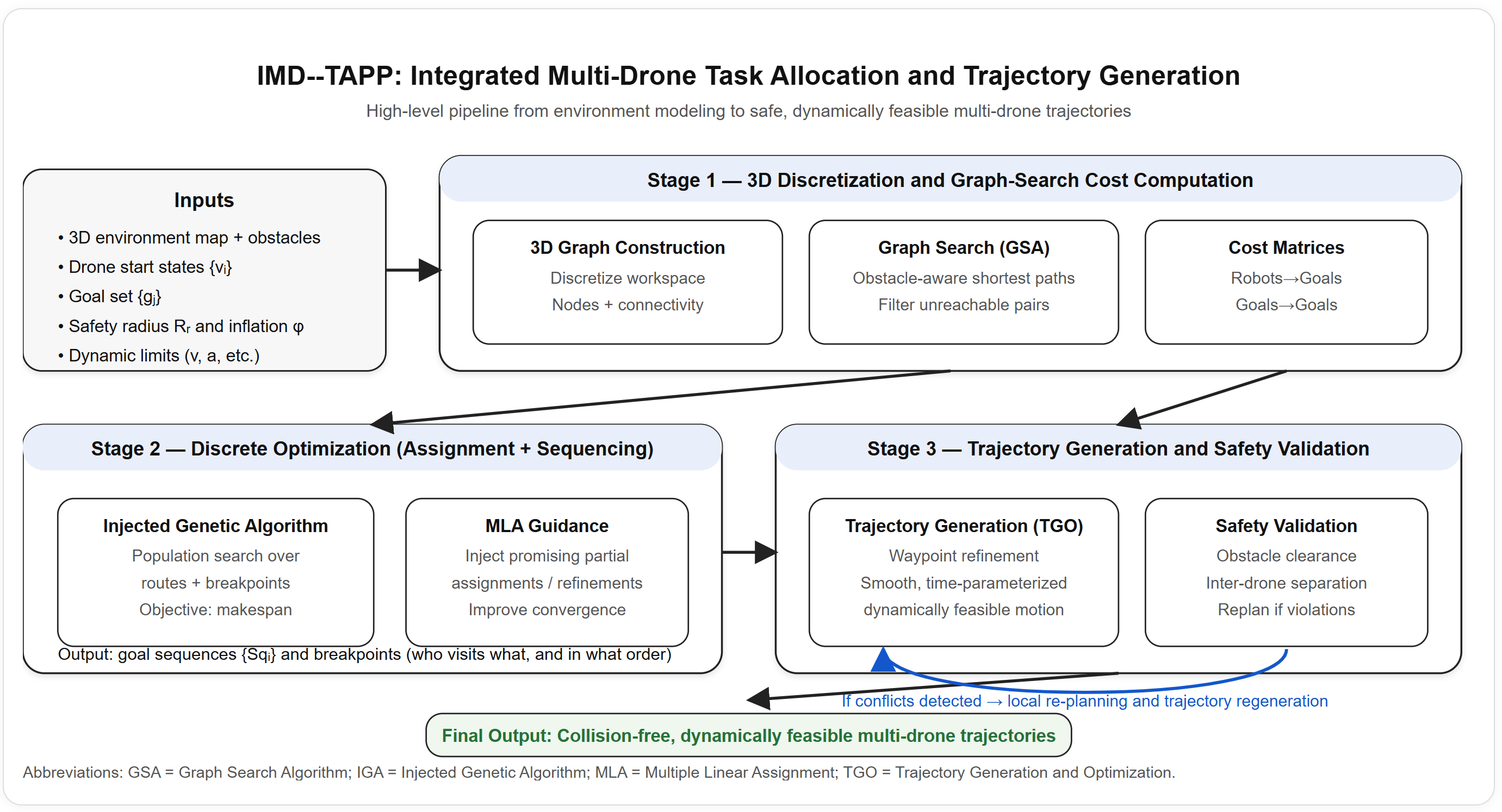}
\caption{High-level overview of the IMD--TAPP pipeline: (i) 3D workspace discretization and graph-search computation of obstacle-aware travel costs,(ii) discrete optimization of goal assignment and visit sequencing using an injected (PSO) algorithm guided by multiple linear assignment, and (iii) smooth trajectory generation followed by safety validation with iterative replanning when necessary.}
\label{fig:framework}
\end{figure}

The proposed pipeline addresses the IMD--TAPP problem by combining (i) 3D graph construction and graph-search-based pathfinding, (ii) IPSO optimization guided by multiple linear assignment, and (iii) trajectory generation and validation for dynamic feasibility and collision avoidance.
Figure~\ref{fig:framework} provides a high-level view of the complete pipeline.

\subsection{Cost-matrix construction via graph search}
The IMD--TAPP discretizes the obstacle-filled workspace into a 3D graph and uses a graph-search algorithm (GSA) to compute feasible paths between relevant pairs of states \cite{arshid2025toward, zhang2025cooperative, alqudsi2024analysis}. Specifically, shortest-path costs are computed between each robot start state and each goal (robots-to-goals matrix), and between each pair of goals (goals-to-goals matrix). Unreachable pairs are filtered to prevent infeasible assignments from entering the optimization stage.

\begin{figure}
\centering
\includegraphics[width=\textwidth]{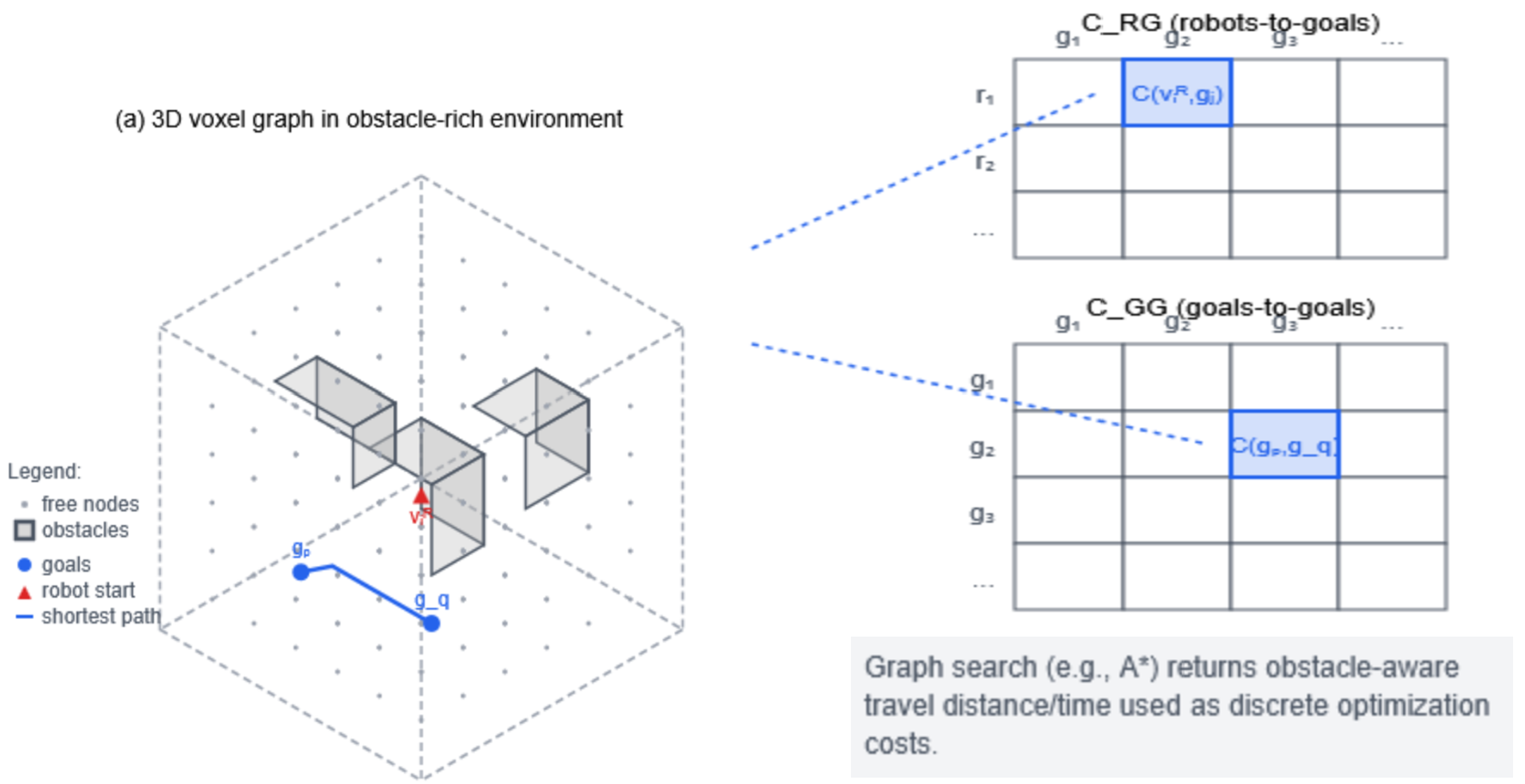}
\caption{3D discretization and obstacle-aware travel-cost computation. (a) The workspace is discretized into a voxel graph with free-space nodes; obstacles block direct paths, requiring graph search to find collision-free routes between robot starts $v_R^i$, goals $g_p$ and $g_q$. (b) Graph-search distances populate the robots-to-goals matrix $\mathbf{C}_{RG}$ (with entries $C(v_R^i, g_j)$) and the goals-to-goals matrix $\mathbf{C}_{GG}$ (with entries $C(g_p, g_q)$), which are then used by the discrete optimizer to determine assignments and visit sequences.}
\label{fig:discretization}
\end{figure}

Figure~\ref{fig:discretization} illustrates the discretization of the 3D workspace into a navigable graph and demonstrates how obstacle-aware shortest paths computed via graph search (e.g., A*) populate the robots-to-goals cost matrix $\mathbf{C}_{RG}$ and goals-to-goals cost matrix $\mathbf{C}_{GG}$ that serve as inputs to the discrete optimization stage.

\begin{figure}
\centering
\resizebox{\linewidth}{!}{%
  \input{fig_cost_encoding.tikz.tex}%
}
\caption{Cost-matrix construction and solution encoding for the discrete optimization stage. Graph search produces robots-to-goals and goals-to-goals travel-cost matrices, which are evaluated through a PSO particle representation encoding a goal-visit permutation and breakpoints that partition the ordered goals among drones while enforcing the visit-once constraint.}
\label{fig:cost-encoding}
\end{figure}
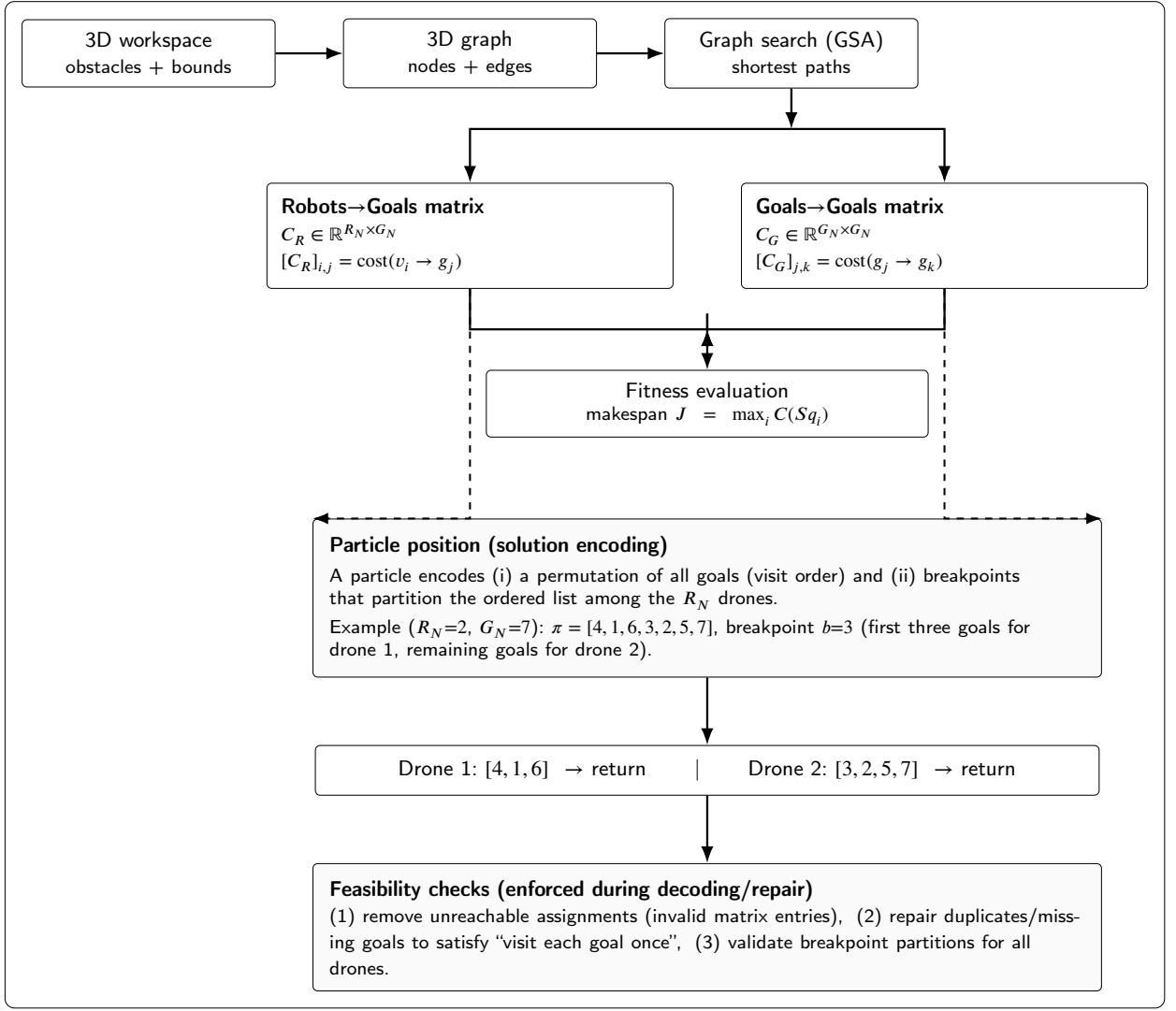

To clarify how environment-aware graph-search costs are converted into a discrete optimization input, Figure~\ref{fig:cost-encoding} illustrates the construction of the robots-to-goals and goals-to-goals cost matrices and the corresponding particle encoding used by the IPSO stage (goal permutation and breakpoints ensuring that each goal is visited exactly once).

\subsection{Assignment and sequencing via injected (PSO) algorithm}
Given travel-cost matrices, the discrete optimization stage seeks an assignment of goals to robots and an ordering of visits that minimizes the makespan objective in \eqref{eq:obj}. The search space is combinatorial, grows rapidly with the number of goals, and is generally NP-hard. The framework therefore adopts a (PSO) algorithm variant \cite{tao2025multi} with injection mechanisms and structured guidance, exploiting both exploration and exploitation to avoid premature convergence. Multiple linear assignment (MLA) is leveraged to generate high-quality candidate sequences and to guide reordering within the PSO evolution loop, improving convergence toward low-cost solutions.

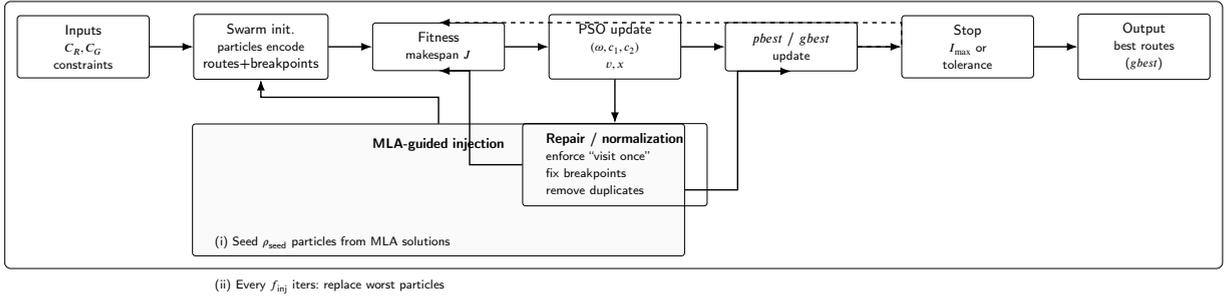
\begin{figure}
\centering
\resizebox{0.98\linewidth}{!}{%
  \input{fig_ipso_mla.tikz.tex}%
}
\caption{Injected PSO with MLA guidance for joint goal assignment and sequencing. Particles encode routes and breakpoints; fitness is evaluated using graph-search costs under the makespan objective. MLA is used both for seeding a fraction of the swarm and for periodic injection (replacement of the worst particles), while a repair step maintains feasibility (each goal visited once and valid partitioning).}
\label{fig:ipso-mla}
\end{figure}

Figure~\ref{fig:ipso-mla} summarizes the IPSO decision loop and highlights where multiple linear assignment (MLA) is used to seed and periodically replace low-quality particles, thereby accelerating convergence while maintaining feasibility under the visit-once constraint.

\subsection{Trajectory generation and safety validation}
After obtaining an optimized discrete plan, each robot's geometric path is converted into a time-parameterized trajectory. The trajectory generation stage uses intermediate waypoints and an optimization routine that promotes smoothness (minimizing snap) while satisfying dynamic constraints \cite{alqudsi2023numerically, lian2024trajectory}. The resulting trajectories are then iteratively validated against the inter-robot separation constraint in \eqref{eq:eq3}; if violations are detected, local re-planning is triggered, and the final trajectories are revalidated \cite{alqudsi2024enhancing}.

\begin{figure}
\centering
\includegraphics[width=\textwidth]{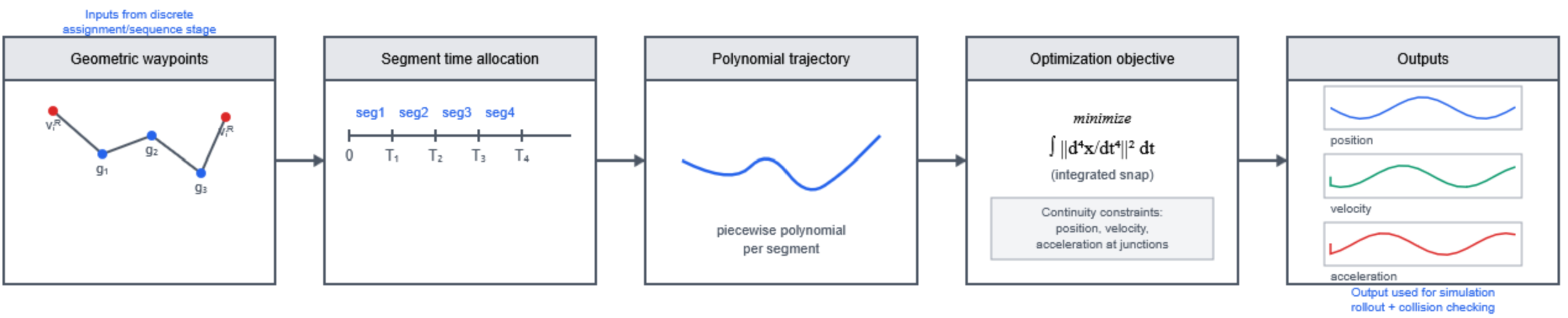}
\caption{Time-parameterized minimum-snap trajectory generation pipeline. The discrete plan provides geometric waypoints (robot start $v_R^i$, assigned goals, and return-to-start). Segment time allocation assigns durations to each path segment. Piecewise polynomial trajectories are then optimized to minimize integrated snap $\int \|\frac{d^4\mathbf{x}}{dt^4}\|^2 dt$ subject to continuity constraints (position, velocity, and acceleration continuous at junctions), producing smooth position, velocity, and acceleration profiles suitable for quadrotor execution.}
\label{fig:traj-pipeline}
\end{figure}

The trajectory generation pipeline is detailed in Figure~\ref{fig:traj-pipeline}, which illustrates how geometric waypoints obtained from the discrete plan are transformed into smooth, dynamically feasible trajectories through time allocation, piecewise polynomial optimization with snap minimization, and enforcement of continuity constraints at waypoint junctions.

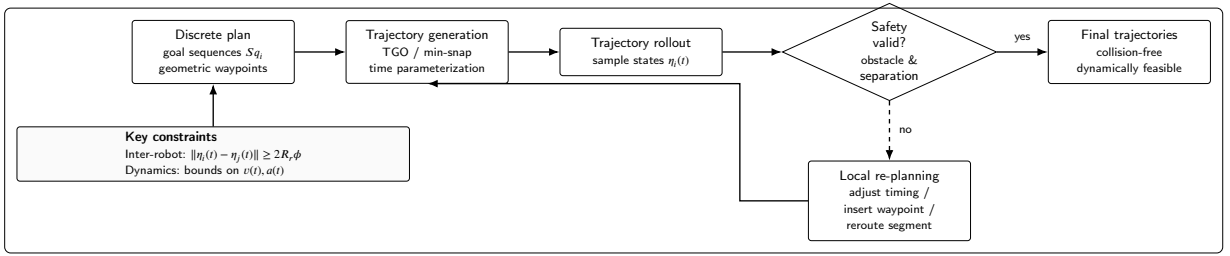
\begin{figure}
\centering
\resizebox{0.98\linewidth}{!}{%
  \input{fig_safety_loop.tikz.tex}%
}
\caption{Safety validation and re-planning loop in the continuous stage. The discrete plan is converted into time-parameterized trajectories; simulated rollouts are checked for obstacle clearance and inter-robot separation (e.g., $\|\eta_i(t)-\eta_j(t)\|\ge 2R_r\phi$). If conflicts occur, local re-planning updates the plan and the trajectories are regenerated until feasibility is achieved.}
\label{fig:safety-loop}
\end{figure}

As shown in Figure~\ref{fig:safety-loop}, the continuous-time stage converts the discrete plan into smooth trajectories and then iteratively checks obstacle clearance and inter-drone separation; if violations are detected, the framework triggers local re-planning and regenerates trajectories until the safety constraints are satisfied.

Algorithm~\ref{algo1} summarizes the complete IMD--TAPP procedure.

\begin{algorithm}[t!]
\SetAlgoLined
\caption{IMD--TAPP Algorithm}\label{algo1}
\KwIn{Number of drones, initial states, goals, safe distance}
\KwOut{Optimized trajectories}
\textbf{Initialization}\;
Input drone numbers, initialize states and goals, define safe distance\;
\textbf{Graph Generation and Pathfinding}\;
Construct 3D graph of environment\;
\For{each drone $i$ and goal $j$}{
    Compute optimal paths using GSA\;
    Populate Robots-to-Goals and Goals-to-Goals cost matrices\;
}
Filter unreachable goals and drones\;
\textbf{Assignment Optimization based on IPSO}\;
\textbf{Initialization}\;
Define IPSO parameters:\;
Swarm size ($N_s$), maximum iterations ($I_{\max}$), inertia weight ($\omega$), cognitive and social coefficients ($c_1, c_2$), velocity bounds ($v_{\max}$)\;
Define injection settings:\;
MLA seeding ratio ($\rho_{\text{seed}}$), MLA replacement frequency ($f_{\text{inj}}$), random perturbation probability ($p_{\text{pert}}$)\;
Define Cost Function (Min-Time / makespan)\;
Initialize particle positions as candidate routes and breakpoints (encode task order + partitioning), seed a subset using MLA solutions\;
Initialize velocities and evaluate fitness for all particles\;
Set personal bests ($pbest$) and global best ($gbest$)\;
\textbf{Swarm Evolution}\;
\For{$t = 1$ to $I_{\max}$}{
    Update velocity and position of each particle using PSO rules\;
    Repair/normalize infeasible encodings (duplicate goals, missing goals, invalid breakpoints)\;
    Evaluate fitness (makespan)\;
    Update $pbest$ and $gbest$\;
    \If{mod($t$, $f_{\text{inj}}$) = 0}{
        Replace the worst particles with MLA-refined solutions (injection)\;
    }
    Apply random perturbation / reinitialization with probability $p_{\text{pert}}$ to avoid stagnation\;
}
Obtain the best routes and breakpoints from $gbest$\;
\textbf{Trajectory Generation}\;
\For{each drone $i$}{
    Compute intermediate waypoints and generate trajectory states based on Trajectory Generation and Optimization (TGO) algorithm \cite{alqudsi2023numerically}\;
    Validate and re-plan for collision avoidance\;
}
Synthesize and re-validate dynamic trajectories \cite{alqudsi2024enhancing}\;
\textbf{Simulation and Visualization}\;
Execute animation and derive simulation results\;
\end{algorithm}

\section{Simulation Setup}\label{sec:setup}
Simulations were executed using MATLAB R2023b on a laptop equipped with an Intel(R) Core(TM) i7-1065G7 CPU @ 1.30GHz, NVIDIA GeForce GTX 1650 GPU with 4~GB dedicated memory, and 16~GB RAM. Quadrotor parameters followed \cite{alqudsi2021robust}. The robots were assumed to have onboard sensing sufficient to detect obstacles and to support replanning within the IMD--TAPP pipeline.

\begin{figure}
\centering
\includegraphics[width=0.8\textwidth]{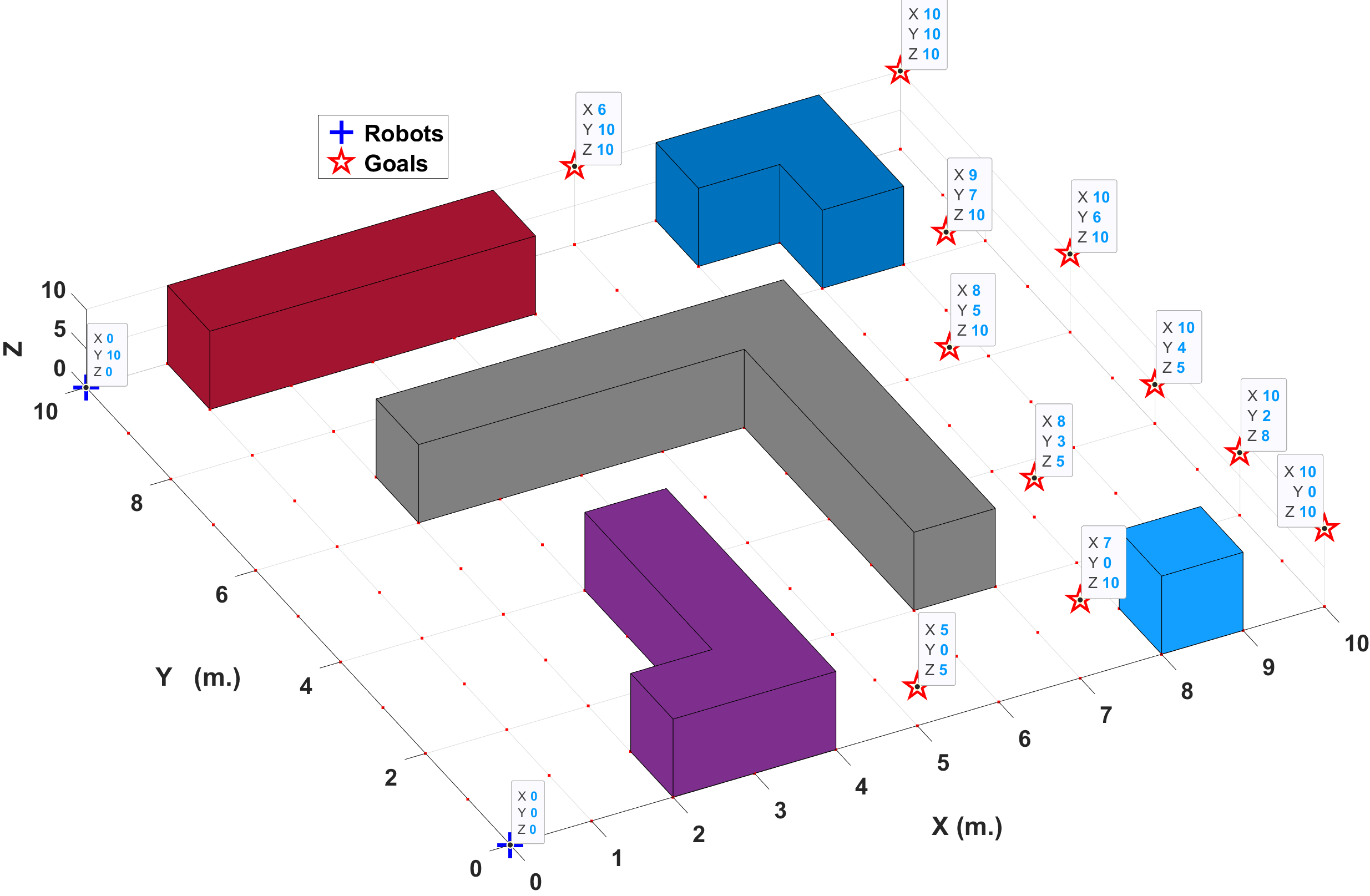}
\caption{3D obstacle-filled environment used in the simulation study, illustrating the initial positions of the robots and the set of goal locations.}\label{fig02}
\end{figure}

Figure~\ref{fig02} shows the representative 3D obstacle-filled environment used to illustrate the algorithm. The initial robot states and goal locations are indicated for reproducibility of the qualitative results. To facilitate reproducibility, Table~\ref{tab:setup} summarizes the main simulation and implementation settings reported in this study.

\begin{table}[t]
\caption{Simulation and implementation settings used in the case study.}
\label{tab:setup}
\centering
\begin{tabular}{ll}
\toprule
Item & Setting \\
\midrule
Software & MATLAB R2023b \\
CPU & Intel Core i7-1065G7 @ 1.30~GHz \\
GPU & NVIDIA GeForce GTX 1650 (4~GB) \\
RAM & 16~GB \\
Robot model parameters & As in \cite{alqudsi2021robust} \\
Environment & 3D workspace with obstacles (Figure~\ref{fig02}) \\
Objective & Minimize makespan (Equation~\ref{eq:obj}) \\
Safety constraint & Inter-robot separation (Equation~\ref{eq:eq3}) \\
\bottomrule
\end{tabular}
\end{table}

\section{Results and Discussion}\label{sec:results}
The IMD--TAPP objective is to determine an optimal sequence of goal visits for each robot such that all goals are serviced exactly once and each robot returns to its start point. The discrete stage therefore balances the combinatorial search over permutations and partitions of goals across robots. As expected for PSO-based methods, the balance between exploration and exploitation affects both convergence speed and solution quality \cite{qi2025multi, liu2025spherical}.

\begin{figure}
\centering
\begin{subfigure}[b]{0.48\textwidth}
    \centering
    \includegraphics[width=\textwidth]{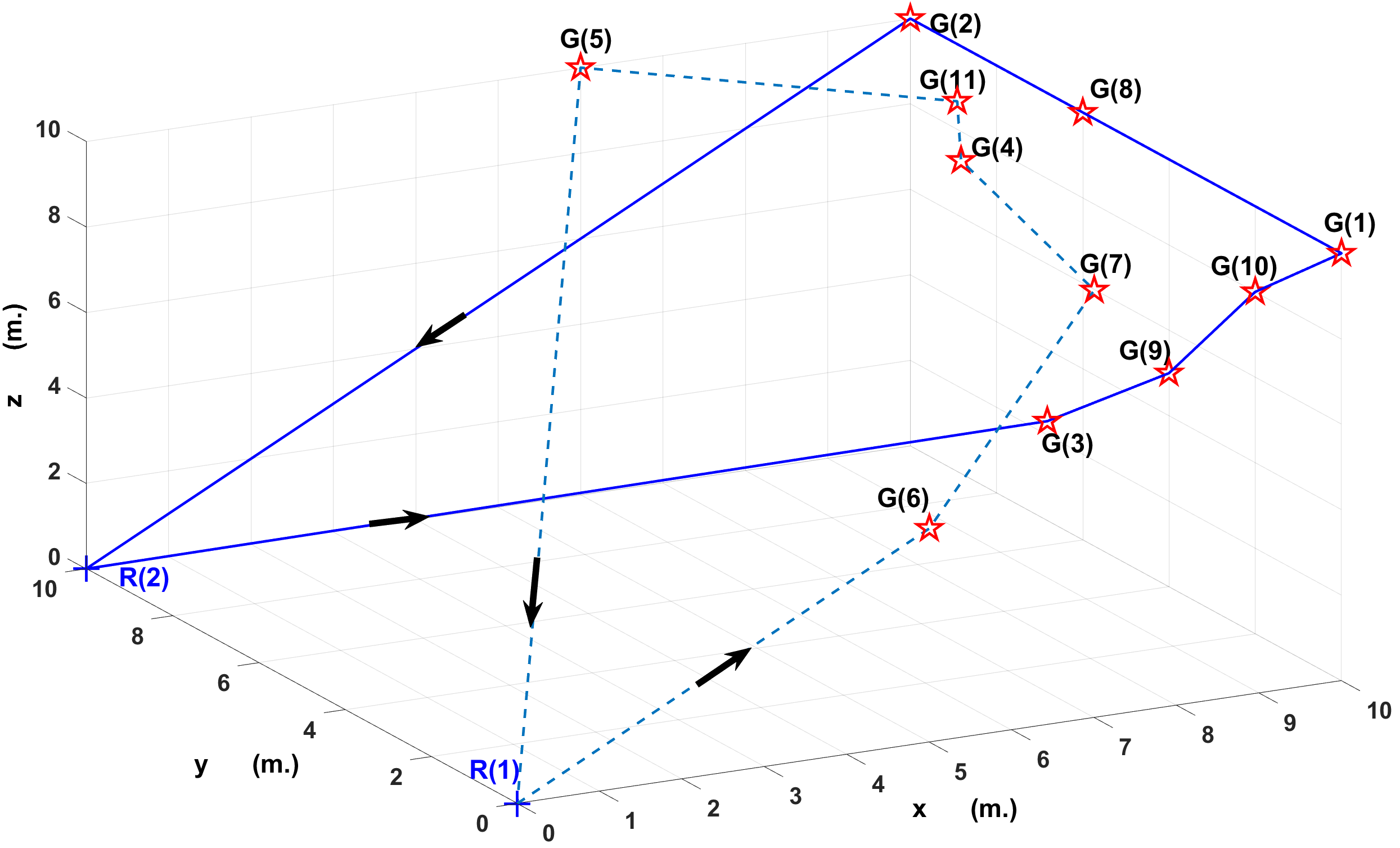}
    \caption{}
    \label{fig:opt-assignment}
\end{subfigure}
\hfill
\begin{subfigure}[b]{0.48\textwidth}
    \centering
    \includegraphics[width=\textwidth]{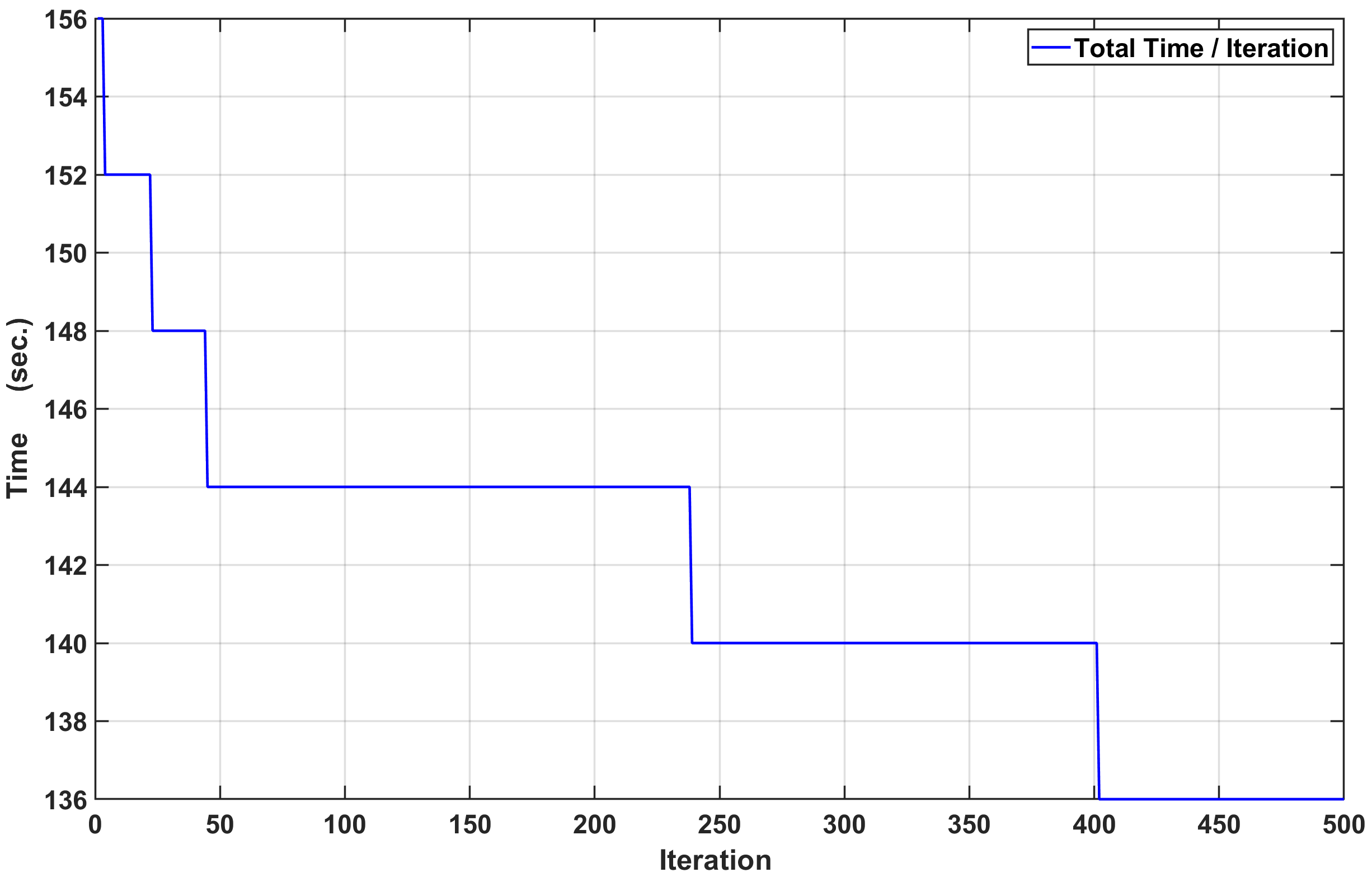}
    \caption{}
    \label{fig:opt-convergence}
\end{subfigure}

\caption{Discrete optimization results. (a) Example optimal sequence and task assignment obtained by the IPSO optimizer, showing the ordered goals assigned to each robot and the required return to the takeoff point. (b) Mission-time objective as a function of iteration number, demonstrating the convergence behavior of the discrete optimization stage.}
\label{fig:optimization-results}
\end{figure}

Following the IMD--TAPP procedure, graph search is first used to generate collision-free geometric paths and to populate cost matrices. The injected (PSO) algorithm then selects routes and breakpoints that minimize the makespan. An example of the resulting optimal sequence and task assignments is provided in Figure~\ref{fig:opt-assignment}, where the assignment indicates which goals are visited by each robot and in what order.

The evolution of the mission-time objective over iterations is shown in Figure~\ref{fig:opt-convergence}. The figure illustrates the convergence behavior of the discrete optimizer toward a low-cost assignment and ordering.

\begin{figure}
\centering

\begin{subfigure}[t]{0.49\textwidth}
  \centering
  \includegraphics[width=\textwidth]{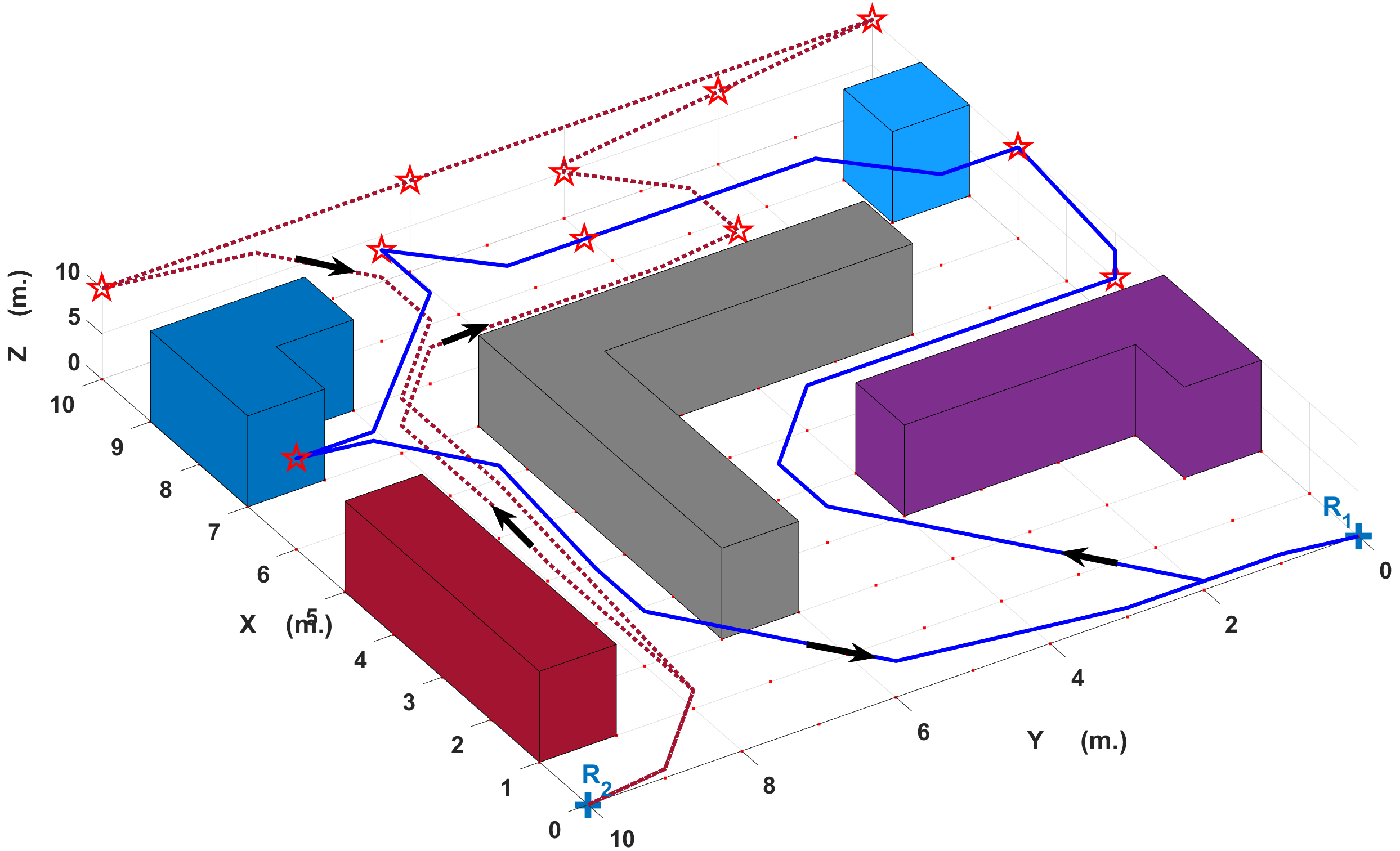}
  \caption{Obstacle-avoiding geometric paths generated for the optimized assignment sequence, illustrating each robot's planned route through its assigned goals.}
  \label{fig05}
\end{subfigure}
\hfill
\begin{subfigure}[t]{0.49\textwidth}
  \centering
  \includegraphics[width=\textwidth]{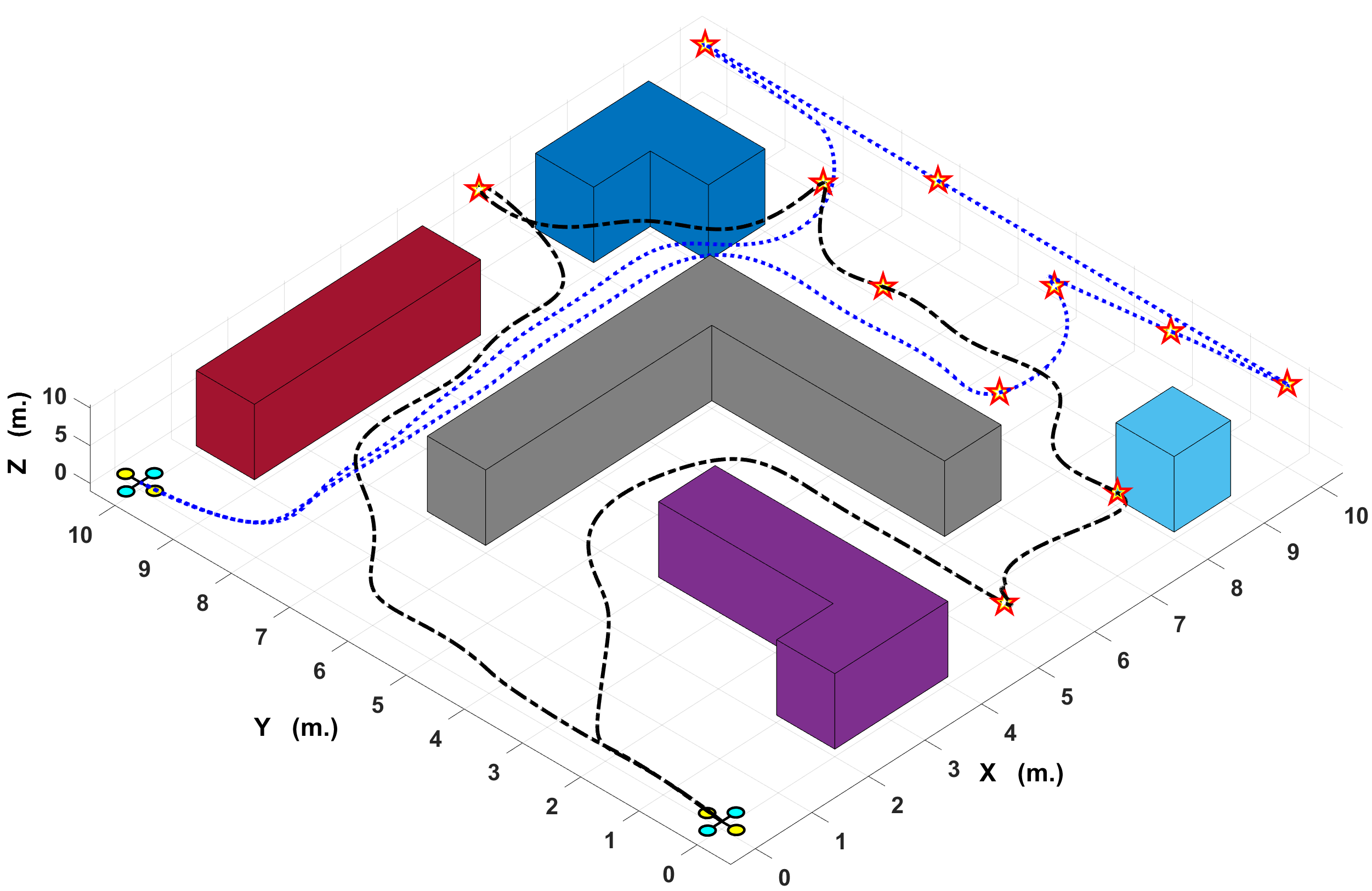}
  \caption{Time-parameterized trajectories generated for each robot, ensuring smooth motion and passage through all designated points while promoting dynamic feasibility via snap minimization.}
  \label{fig06}
\end{subfigure}

\caption{Comparison of (a) geometric paths and (b) time-parameterized trajectories for the optimized assignment sequence.}
\label{fig:paths_and_trajs}
\end{figure}

After selecting the optimal discrete plan, geometric paths are generated for each robot and visualized in Figure~\ref{fig05}. These paths provide obstacle-avoiding routes connecting each robot's start state to its assigned goals.

Next, time-parameterized trajectories are synthesized from the geometric paths. Figure~\ref{fig06} presents the resulting trajectories, which satisfy waypoint passage and promote smooth motion by minimizing snap, while respecting feasibility constraints. In this representative case study with two quadrotor robots, the minimum total mission time achieved by the algorithm was 136~sec.

\begin{figure}
\centering

\begin{subfigure}[t]{0.4\textwidth}
  \centering
  \includegraphics[width=\textwidth]{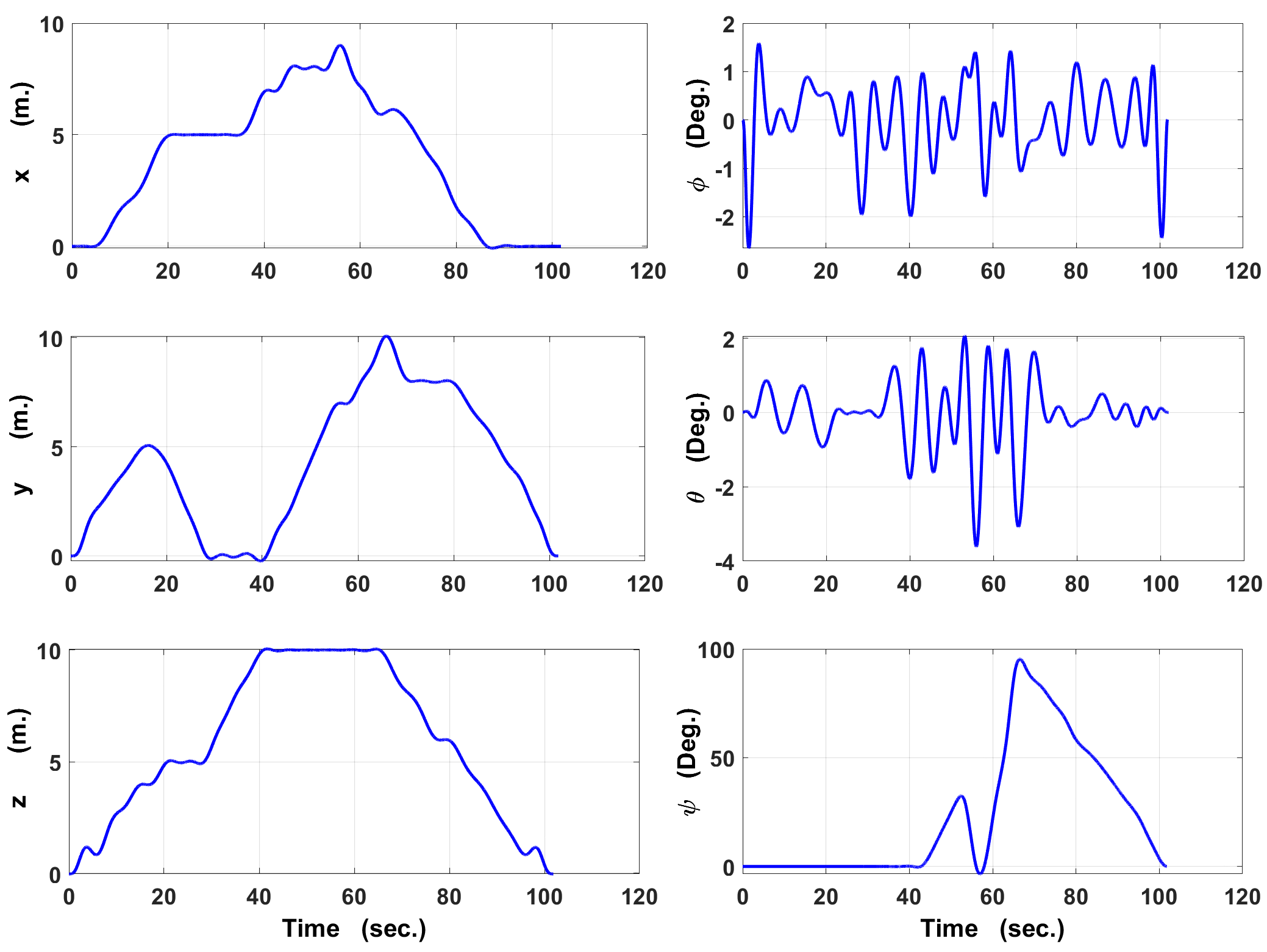}
  \caption{Sample simulation results showing the states of each robot as a function of time for the generated time-parameterized trajectories.}
  \label{fig07}
\end{subfigure}
\hfill
\begin{subfigure}[t]{0.58\textwidth}
  \centering
  \includegraphics[width=\textwidth]{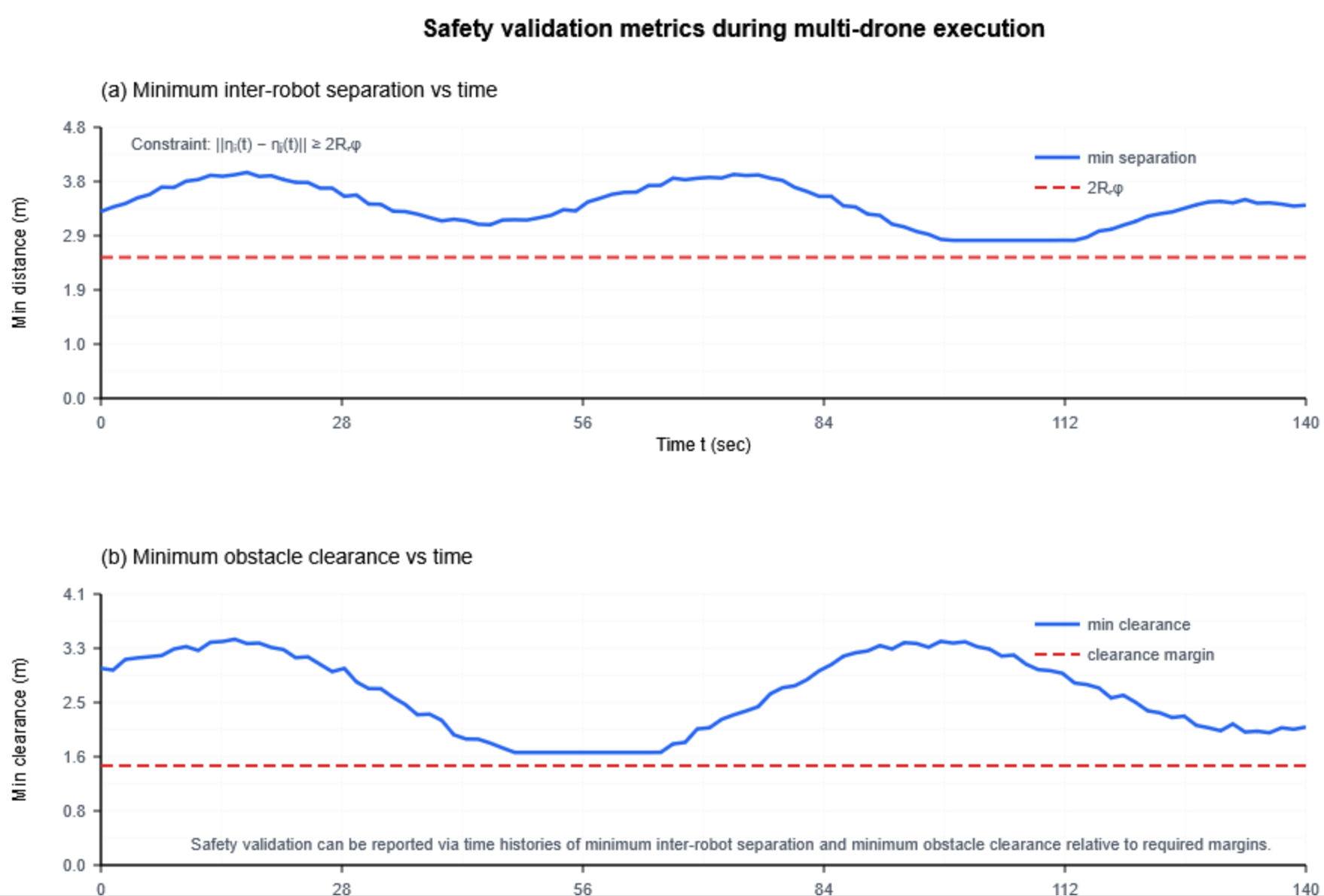}
  \caption{Safety validation metrics during multi-drone execution. (i) Minimum inter-robot separation versus time, demonstrating that the constraint is satisfied throughout the mission. (i) Minimum obstacle clearance versus time, showing that all robots maintain adequate separation from obstacles. Both metrics remain above their safety thresholds, validating the collision-free operation of the generated trajectories.}
  \label{fig:safety-metrics}
\end{subfigure}

\caption{Simulation outputs: (a) robot states over time and (b) safety validation metrics for the executed trajectories.}
\label{fig:sim-and-safety}
\end{figure}

For a detailed view of the robot states and their evolution over time, Figure~\ref{fig07} provides sample state trajectories. Overall, the results demonstrate that IMD--TAPP can allocate tasks and generate safe trajectories for aerial robot teams operating in cluttered 3D environments.

To validate that the generated trajectories satisfy the required safety constraints throughout a mission, Figure~\ref{fig:safety-metrics} presents time histories of the minimum inter-robot separation and minimum obstacle clearance of another mission scenario. Both metrics remain above their respective safety thresholds for the entire mission duration, confirming collision-free operation.

\section{Conclusion and Future Work}
\label{sec:conclusion}
This paper presented IMD--TAPP, an integrated framework for multi-drone goal allocation, visit sequencing, and trajectory generation in obstacle-rich 3D environments. The framework (i) computes obstacle-aware travel costs via 3D graph search, (ii) solves the coupled assignment-and-ordering problem using IPSO with MLA guidance under a makespan objective, and (iii) produces smooth minimum-snap trajectories with iterative safety validation and local re-planning. MATLAB simulations demonstrate end-to-end feasibility on representative 3D obstacle fields, including a case study in which two drones service multiple goals and complete the mission in 136~sec while satisfying obstacle-avoidance and inter-robot separation constraints.

Future work should strengthen the empirical evaluation and broaden applicability. Hardware experiments are needed to evaluate robustness to state-estimation uncertainty, communication delay, and model mismatch, and to validate computational performance under real-time constraints. Finally, extending the framework to handle heterogeneous robots, limited battery budgets, and mission-level constraints (e.g., time windows, precedence, or revisits) would further increase its relevance to practical inspection, monitoring, and search-and-rescue deployments.

\section*{Declaration of competing interest}
The authors declare that they have no known competing financial interests or personal relationships that could have appeared to influence the work reported in this paper.

\bibliographystyle{model1-num-names} 

\bibliography{cas-refs2} 

\end{document}

%% file: fig_cost_encoding.tikz.tex
\begin{tikzpicture}[
  font=\small,
  >=Latex,
  box/.style={draw, rounded corners=2pt, align=center, inner sep=5pt, text width=34mm},
  mat/.style={draw, rounded corners=2pt, align=left, inner sep=6pt, text width=56mm},
  wide/.style={draw, rounded corners=2pt, align=center, inner sep=6pt, text width=112mm},
  panel/.style={draw, rounded corners=2pt, fill=black!2, align=left, inner sep=7pt, text width=112mm},
  arrow/.style={-Latex, line width=0.9pt},
  dashedarrow/.style={-Latex, dashed, line width=0.8pt}
]
\usetikzlibrary{positioning,calc,fit}

\node[box] (env) {3D workspace\\\footnotesize obstacles + bounds};
\node[box, right=10mm of env] (graph) {3D graph\\\footnotesize nodes + edges};
\node[box, right=10mm of graph] (gsa) {Graph search (GSA)\\\footnotesize shortest paths};

\draw[arrow] (env) -- (graph);
\draw[arrow] (graph) -- (gsa);

\node[mat, below=14mm of graph] (rg) {%
\textbf{Robots$\rightarrow$Goals matrix}\\[0.8mm]
\footnotesize $C_R \in \mathbb{R}^{R_N \times G_N}$\\
\footnotesize $[C_R]_{i,j}= \mathrm{cost}(v_i \rightarrow g_j)$};

\node[mat, right=10mm of rg] (gg) {%
\textbf{Goals$\rightarrow$Goals matrix}\\[0.8mm]
\footnotesize $C_G \in \mathbb{R}^{G_N \times G_N}$\\
\footnotesize $[C_G]_{j,k}= \mathrm{cost}(g_j \rightarrow g_k)$};

\coordinate (Jgsa) at ($(gsa.south)+(0,-6mm)$);

\draw[arrow] (gsa.south) -- (Jgsa);
\draw[arrow] (Jgsa) -| (rg.north);
\draw[arrow] (Jgsa) -| (gg.north);

\node[box, below=12mm of $(rg.south)!0.5!(gg.south)$, text width=62mm] (fitness) {%
Fitness evaluation\\[-0.5mm]
\footnotesize makespan $J=\max_{i} C(Sq_i)$};


\path (rg.south) -- (gg.south) coordinate[midway] (Mmat);
\coordinate (Jmat) at ([yshift=-6mm]Mmat);

\draw[arrow] (rg.south) -- ++(0,-6mm) -| (Jmat);
\draw[arrow] (gg.south) -- ++(0,-6mm) -| (Jmat);
\draw[arrow] (Jmat) -- (fitness.north);

\node[panel, below=12mm of fitness] (enc) {%
\begin{minipage}[t]{108mm}
\raggedright
\textbf{Particle position (solution encoding)}\\[1.2mm]
\footnotesize
A particle encodes (i) a permutation of all goals (visit order) and (ii) breakpoints that partition the ordered list among the $R_N$ drones.\\[0.9mm]
Example ($R_N{=}2$, $G_N{=}7$): $\pi=[4,1,6,3,2,5,7]$, breakpoint $b{=}3$ (first three goals for drone~1, remaining goals for drone~2).
\end{minipage}
};

\draw[dashedarrow] (rg.south) |- (enc.north west);
\draw[dashedarrow] (gg.south) |- (enc.north east);

\node[wide, below=10mm of enc] (split) {%
Drone 1: $[4,1,6]\rightarrow$ return \hspace{6mm}|\hspace{6mm}
Drone 2: $[3,2,5,7]\rightarrow$ return};

\draw[arrow] (enc.south) -- (split.north);

\node[panel, below=10mm of split] (feas) {%
\textbf{Feasibility checks (enforced during decoding/repair)}\\
\footnotesize (1) remove unreachable assignments (invalid matrix entries),\;
(2) repair duplicates/missing goals to satisfy ``visit each goal once'',\;
(3) validate breakpoint partitions for all drones.};

\draw[arrow] (split.south) -- (feas.north);

\node[draw, rounded corners=4pt,
      fit=(env)(gsa)(rg)(gg)(fitness)(enc)(split)(feas),
      inner sep=7pt] (frame) {};
\end{tikzpicture}

%% file: fig_ipso_mla.tikz.tex
\begin{tikzpicture}[
  font=\small,
  block/.style={draw, rounded corners=2pt, align=center, inner sep=6pt, minimum width=30mm},
  sub/.style={draw, rounded corners=2pt, align=left, inner sep=6pt, minimum width=42mm},
  arrow/.style={-Latex, line width=0.9pt},
  loop/.style={-Latex, line width=0.9pt, dashed},
  soft/.style={fill=black!2, draw, rounded corners=2pt}
]
\usetikzlibrary{positioning,arrows.meta,fit,calc}

\node[block] (inputs) {Inputs\\\footnotesize $C_R, C_G$\\\footnotesize constraints};
\node[block, right=10mm of inputs] (init) {Swarm init.\\\footnotesize particles encode\\routes+breakpoints};
\node[block, right=10mm of init] (eval) {Fitness\\\footnotesize makespan $J$};
\node[block, right=10mm of eval] (update) {PSO update\\\footnotesize $(\omega,c_1,c_2)$\\\footnotesize $v,x$};
\node[block, right=10mm of update] (best) {$pbest$ / $gbest$\\\footnotesize update};

\draw[arrow] (inputs) -- (init);
\draw[arrow] (init) -- (eval);
\draw[arrow] (eval) -- (update);
\draw[arrow] (update) -- (best);

\node[soft, below=12mm of eval, minimum width=112mm, minimum height=30mm, anchor=north] (inj) {};
\node[anchor=north] at ($(inj.north)+(0,-2mm)$) {\textbf{MLA-guided injection}};
\node[anchor=west] at ($(inj.west)+(4mm,-12mm)$) {\footnotesize (i) Seed $\rho_{\text{seed}}$ particles from MLA solutions};
\node[anchor=west] at ($(inj.west)+(4mm,-22mm)$) {\footnotesize (ii) Every $f_{\text{inj}}$ iters: replace worst particles};

\node[sub, below=10mm of update] (repair) {\textbf{Repair / normalization}\\
\footnotesize enforce ``visit once''\\
\footnotesize fix breakpoints\\
\footnotesize remove duplicates};

\draw[arrow] (update.south) -- (repair.north);
\draw[arrow] (repair.west) -- ++(-12mm,0) |- (eval.south);

\draw[loop] (best.east) -- ++(10mm,0) |- (eval.north);

\draw[arrow] (inj.north) -- ++(0,6mm) -| (init.south);
\draw[arrow] (inj.east) -- ++(12mm,0) |- (best.south);

\node[block, right=10mm of best] (term) {Stop\\\footnotesize $I_{\max}$ or\\\footnotesize tolerance};
\node[block, right=10mm of term] (out) {Output\\\footnotesize best routes\\\footnotesize ($gbest$)};

\draw[arrow] (best) -- (term);
\draw[arrow] (term) -- (out);

\node[draw, rounded corners=4pt, fit=(inputs)(out)(inj)(repair), inner sep=8pt] (frame) {};
\end{tikzpicture}

%% file: fig_safety_loop.tikz.tex
\begin{tikzpicture}[
  font=\small,
  block/.style={draw, rounded corners=2pt, align=center, inner sep=6pt, minimum width=38mm},
  decision/.style={draw, diamond, aspect=2.2, inner sep=2pt, align=center},
  arrow/.style={-Latex, line width=0.9pt},
  dashedarrow/.style={-Latex, line width=0.9pt, dashed},
  soft/.style={fill=black!2, draw, rounded corners=2pt}
]
\usetikzlibrary{positioning,arrows.meta,fit,calc}

\node[block] (disc) {Discrete plan\\\footnotesize goal sequences $Sq_i$\\\footnotesize geometric waypoints};
\node[block, right=12mm of disc] (tgo) {Trajectory generation\\\footnotesize TGO / min-snap\\\footnotesize time parameterization};
\node[block, right=12mm of tgo] (sim) {Trajectory rollout\\\footnotesize sample states $\eta_i(t)$};
\node[decision, right=14mm of sim] (check) {Safety\\valid?\\\footnotesize obstacle \&\\separation};

\node[block, below=14mm of check] (replan) {Local re-planning\\\footnotesize adjust timing /\\\footnotesize insert waypoint /\\\footnotesize reroute segment};

\node[block, right=12mm of check] (final) {Final trajectories\\\footnotesize collision-free\\\footnotesize dynamically feasible};

\draw[arrow] (disc) -- (tgo);
\draw[arrow] (tgo) -- (sim);
\draw[arrow] (sim) -- (check);
\draw[arrow] (check) -- node[above, yshift=1mm]{\footnotesize yes} (final);
\draw[dashedarrow] (check) -- node[right, xshift=1mm]{\footnotesize no} (replan);
\draw[arrow] (replan.west) -- ++(-16mm,0) |- (tgo.south);

\node[soft, below=10mm of disc, minimum width=92mm, align=left] (constr) {\textbf{Key constraints}\\
\footnotesize Inter-robot: $\|\eta_i(t)-\eta_j(t)\|\ge 2R_r\phi$\\
\footnotesize Dynamics: bounds on $v(t), a(t)$};

\draw[arrow] (constr.north) -- (disc.south);

\node[draw, rounded corners=4pt, fit=(disc)(final)(replan)(constr), inner sep=8pt] (frame) {};
\end{tikzpicture}

%% file: cas-refs2.bib
@article{rojas2021unmanned,
  title={Unmanned aerial vehicles/drones in vehicle routing problems: a literature review},
  author={Rojas Viloria, Daniela and Solano-Charris, Elyn L and Mu{\~n}oz-Villamizar, Andr{\'e}s and Montoya-Torres, Jairo R},
  journal={International Transactions in Operational Research},
  volume={28},
  number={4},
  pages={1626--1657},
  year={2021},
  publisher={Wiley Online Library}
}

@article{du2025survey,
  title={A Survey on Autonomous and Intelligent Swarms of Uncrewed Aerial Vehicles (UAVs)},
  author={Du, Zhenpeng and Luo, Chunbo and Min, Geyong and Wu, Jia and Luo, Cai and Pu, Jian and Li, Shuai},
  journal={IEEE Transactions on Intelligent Transportation Systems},
  year={2025},
  publisher={IEEE}
}

@article{dai2025heterogeneous,
  title={Heterogeneous multi-robot task allocation and scheduling via reinforcement learning},
  author={Dai, Weiheng and Rai, Utkarsh and Chiun, Jimmy and Yuhong, Cao and Sartoretti, Guillaume},
  journal={IEEE Robotics and Automation Letters},
  year={2025},
  publisher={IEEE}
}

@inproceedings{alqudsi2025advancements,
  title={Advancements and Challenges in VTOL UAVs Configurations and Emerging Trends},
  author={Alqudsi, Yunes and Sulaiman, Husam},
  booktitle={2025 5th International Conference on Emerging Smart Technologies and Applications (eSmarTA)},
  pages={1--8},
  year={2025},
  organization={IEEE}
}

@article{song2025comparative,
  title={Comparative Analysis of Centralized and Distributed Multi-UAV Task Allocation Algorithms: A Unified Evaluation Framework},
  author={Song, Yunze and Ma, Zhexuan and Chen, Nuo and Zhou, Shenghao and Srigrarom, Sutthiphong},
  journal={Drones},
  volume={9},
  number={8},
  pages={530},
  year={2025},
  publisher={MDPI}
}

@article{ref2,
  author  = {del Cerro, J. and Cruz Ulloa, C. and Barrientos, A. and de Le{\'o}n Rivas, J.},
  title   = {Unmanned aerial vehicles in agriculture: A survey},
  journal = {Agronomy},
  volume  = {11},
  number  = {2},
  pages   = {203},
  year    = {2021}
}

@article{alqudsi2025injected,
  title={An injected multi-objective metaheuristic approach for optimizing aerial-robot swarm guidance in cluttered environments},
  author={Alqudsi, Yunes},
  journal={Applied Soft Computing},
  pages={113379},
  year={2025},
  publisher={Elsevier}
}

@article{alqudsi2025towards,
  title={Towards optimal guidance of autonomous swarm drones in dynamic constrained environments},
  author={Alqudsi, Yunes and Makaraci, Murat},
  journal={Expert Systems},
  volume={42},
  number={6},
  pages={e70067},
  year={2025},
  publisher={Wiley Online Library}
}

@misc{ref3,
  author = {Turpin, M.},
  title  = {Safe, scalable, and complete motion planning of large teams of interchangeable robots},
  year   = {2014},
  note   = {Publicly Accessible Penn Dissertations}
}

@inproceedings{ref4,
  author    = {Zhang, Y. and Zhang, L. and Wang, H. and Bustamante, F. E. and Rubenstein, M.},
  title     = {SwarmTalk-towards benchmark software suites for swarm robotics platforms},
  booktitle = {Proceedings of the 19th International Conference on Autonomous Agents and MultiAgent Systems},
  pages     = {1638--1646},
  year      = {2020}
}

@inproceedings{alqudsi2024coordinated,
  title={Coordinated formation control for swarm flying robots},
  author={Alqudsi, Yunes},
  booktitle={2024 1st International Conference on Emerging Technologies for Dependable Internet of Things (ICETI)},
  pages={1--8},
  year={2024},
  organization={IEEE}
}

@article{maity2023flying,
  author  = {Maity, Ritu and Mishra, Ruby and Pattnaik, Prasant Kumar},
  title   = {Flying robot path planning techniques and its trends},
  journal = {Materials Today: Proceedings},
  volume  = {80},
  pages   = {2187--2192},
  year    = {2023},
  publisher = {Elsevier}
}

@article{javed2024state,
  author  = {Javed, Sadaf and Hassan, Ali and Ahmad, Rizwan and Ahmed, Waqas and Ahmed, Rehan and Saadat, Ahsan and Guizani, Mohsen},
  title   = {State-of-the-Art and Future Research Challenges in UAV Swarms},
  journal = {IEEE Internet of Things Journal},
  year    = {2024},
  publisher = {IEEE}
}

@article{alqudsi2025integrated,
  author  = {Alqudsi, Y.},
  title   = {Integrated Optimization of Simultaneous Target Assignment and Path Planning for Aerial Robot Swarm},
  journal = {The Journal of Supercomputing},
  volume  = {81},
  number  = {1},
  pages   = {1--24},
  year    = {2025},
  publisher = {Springer}
}

@article{shorinwa2024distributed,
  author  = {Shorinwa, Ola and Halsted, Trevor and Yu, Javier and Schwager, Mac},
  title   = {Distributed Optimization Methods for Multi-robot Systems: Part 1---A Tutorial},
  journal = {IEEE Robotics {\&} Automation Magazine},
  year    = {2024},
  publisher = {IEEE}
}

@article{liu2025spherical,
  title={A spherical vector-based adaptive evolutionary particle swarm optimization for UAV path planning under threat conditions},
  author={Liu, Yanfei and Zhang, Hao and Zheng, Hao and Li, Qi and Tian, Qi},
  journal={Scientific Reports},
  volume={15},
  number={1},
  pages={2116},
  year={2025},
  publisher={Nature Publishing Group UK London}
}

@article{qi2025multi,
  title={Multi-UAV path planning considering multiple energy consumptions via an improved bee foraging learning particle swarm optimization algorithm},
  author={Qi, Yuanhang and Jiang, Haoran and Huang, Gewen and Yang, Liang and Wang, Fujie and Xu, Yunjian},
  journal={Scientific Reports},
  volume={15},
  number={1},
  pages={1--16},
  year={2025},
  publisher={Nature Publishing Group}
}

@article{lian2024trajectory,
  title={Trajectory optimization of unmanned surface vehicle based on improved minimum snap},
  author={Lian, Lian and Zong, Xuejun and He, Kan and Yang, Zhongjun},
  journal={Ocean Engineering},
  volume={302},
  pages={117719},
  year={2024},
  publisher={Elsevier}
}

@article{chen2024distributed,
  title={Distributed Task Allocation for Multiple UAVs Based on Swarm Benefit Optimization},
  author={Chen, Yiting and Chen, Runfeng and Huang, Yuchong and Xiong, Zehao and Li, Jie},
  journal={Drones},
  volume={8},
  number={12},
  pages={766},
  year={2024},
  publisher={MDPI}
}

@article{ghauri2024review,
  title={A review of multi-uav task allocation algorithms for a search and rescue scenario},
  author={Ghauri, Sajjad A and Sarfraz, Mubashar and Qamar, Rahim Ali and Sohail, Muhammad Farhan and Khan, Sheraz Alam},
  journal={Journal of Sensor and Actuator Networks},
  volume={13},
  number={5},
  pages={47},
  year={2024},
  publisher={MDPI}
}

@article{wang2024decentralized,
  author  = {Wang, Ziquan and Li, Juan and Li, Jie and Liu, Chang},
  title   = {A decentralized decision-making algorithm of UAV swarm with information fusion strategy},
  journal = {Expert Systems with Applications},
  volume  = {237},
  pages   = {121444},
  year    = {2024},
  publisher = {Elsevier}
}

@article{peng2021review,
  title={Review of dynamic task allocation methods for UAV swarms oriented to ground targets},
  author={Peng, Qiang and Wu, Husheng and Xue, Ruisong},
  journal={Complex System Modeling and Simulation},
  volume={1},
  number={3},
  pages={163--175},
  year={2021},
  publisher={TUP}
}

@inproceedings{alqudsi2024analysis,
  title={Analysis and implementation of motion planning algorithms for real-time navigation of aerial robots in dynamic environments},
  author={Alqudsi, Yunes},
  booktitle={2024 4th International Conference on Emerging Smart Technologies and Applications (eSmarTA)},
  pages={1--10},
  year={2024},
  organization={IEEE}
}

@article{tao2025multi,
  title={Multi-Strategy Improved Particle Swarm Optimization Algorithm for Path Planning of UAV in 3-D Low Altitude Urban Environment},
  author={Tao, Fazhan and Chen, Zezheng and Wang, Zhikai and Zhu, Longlong and Wang, Jun},
  journal={IEEE Internet of Things Journal},
  year={2025},
  publisher={IEEE}
}

@article{ghanem2025decision,
  title={A decision support framework on simulation fidelity for transferable and autonomously optimised swarm behaviour},
  author={Ghanem, Reda and Ali, Ismail M and Kasmarik, Kathryn and Garratt, Matthew},
  journal={International Journal of Production Research},
  pages={1--22},
  year={2025},
  publisher={Taylor \& Francis}
}

@article{almansoori2024evolution,
  author  = {Almansoori, Ahmed and Alkilabi, Muhanad and Tuci, Elio},
  title   = {On the evolution of adaptable and scalable mechanisms for collective decision-making in a swarm of robots},
  journal = {Swarm Intelligence},
  volume  = {18},
  number  = {1},
  pages   = {79--99},
  year    = {2024},
  publisher = {Springer}
}

@article{du2023multi,
  author  = {Du, Yuwen},
  title   = {Multi-UAV Search and Rescue with Enhanced A* Algorithm Path Planning in 3D Environment},
  journal = {International Journal of Aerospace Engineering},
  volume  = {2023},
  number  = {1},
  pages   = {8614117},
  year    = {2023},
  publisher = {Wiley}
}

@article{yao2020integrated,
  author  = {Yao, Zhihong and Jiang, Haoran and Cheng, Yang and Jiang, Yangsheng and Ran, Bin},
  title   = {Integrated schedule and trajectory optimization for connected automated vehicles in a conflict zone},
  journal = {IEEE Transactions on Intelligent Transportation Systems},
  volume  = {23},
  number  = {3},
  pages   = {1841--1851},
  year    = {2020},
  publisher = {IEEE}
}

@inproceedings{meng2007hybrid,
  author       = {Meng, Yan and Kazeem, Olorundamilola and Muller, Juan C.},
  title        = {A hybrid ACO/PSO control algorithm for distributed swarm robots},
  booktitle    = {2007 IEEE Swarm Intelligence Symposium},
  pages        = {273--280},
  year         = {2007},
  organization = {IEEE}
}

@article{ref16,
  author  = {Shuai, Y. and Yunfeng, S. and Kai, Z.},
  title   = {An effective method for solving multiple travelling salesman problem based on NSGA-II},
  journal = {Systems Science {\&} Control Engineering},
  volume  = {7},
  number  = {2},
  pages   = {108--116},
  year    = {2019}
}

@article{li2023dynamic,
  author  = {Li, Yongqi and Li, Shengquan and Zhang, Yumei and Zhang, Weidong and Lu, Haibo},
  title   = {Dynamic route planning for a USV-UAV multi-robot system in the rendezvous task with obstacles},
  journal = {Journal of Intelligent {\&} Robotic Systems},
  volume  = {107},
  number  = {4},
  pages   = {52},
  year    = {2023},
  publisher = {Springer}
}

@article{burger2013complete,
  author  = {Burger, Mernout and Huiskamp, Marco and Keviczky, Tam{\'a}s},
  title   = {Complete field coverage as a multi-vehicle routing problem},
  journal = {IFAC Proceedings Volumes},
  volume  = {46},
  number  = {18},
  pages   = {97--102},
  year    = {2013},
  publisher = {Elsevier}
}

@article{osaba2020soft,
  author  = {Osaba, Eneko and Del Ser, Javier and Iglesias, Andres and Yang, Xin-She},
  title   = {Soft computing for swarm robotics: new trends and applications},
  journal = {Journal of Computational Science},
  volume  = {39},
  pages   = {101049},
  year    = {2020},
  publisher = {Elsevier}
}

@inproceedings{ref8,
  author    = {Saad, S. and Wan Jaafar, W. N. and Jamil, S. J.},
  title     = {Solving standard traveling salesman problem and multiple traveling salesman problem by using branch-and-bound},
  booktitle = {AIP Conference Proceedings},
  pages     = {1406--1411},
  year      = {2013},
  publisher = {American Institute of Physics}
}

@article{alqudsi2024enhancing,
  author  = {Alqudsi, Y. S. and Saleh, R. A. A. and Makaraci, M. and Ertun{\c{c}}, H. M.},
  title   = {Enhancing aerial robots performance through robust hybrid control and metaheuristic optimization of controller parameters},
  journal = {Neural Computing and Applications},
  volume  = {36},
  number  = {1},
  pages   = {413--424},
  year    = {2024},
  publisher = {Springer}
}

@article{rezaee2024comprehensive,
  author  = {Rezaee, M. R. and Hamid, N. A. W. A. and Hussin, M. and Zukarnain, Z. A.},
  title   = {Comprehensive Review of Drones Collision Avoidance Schemes: Challenges and Open Issues},
  journal = {IEEE Transactions on Intelligent Transportation Systems},
  year    = {2024},
  publisher = {IEEE}
}

@article{saunders2024autonomous,
  author  = {Saunders, J. and Saeedi, S. and Li, W.},
  title   = {Autonomous aerial robotics for package delivery: A technical review},
  journal = {Journal of Field Robotics},
  volume  = {41},
  number  = {1},
  pages   = {3--49},
  year    = {2024},
  publisher = {Wiley}
}

@article{alqudsi2023numerically,
  author  = {Alqudsi, Y. and Makaraci, M. and Kassem, A. and El-Bayoumi, G.},
  title   = {A numerically-stable trajectory generation and optimization algorithm for autonomous quadrotor UAVs},
  journal = {Robotics and Autonomous Systems},
  volume  = {170},
  pages   = {104532},
  year    = {2023},
  publisher = {Elsevier}
}

@inproceedings{alqudsi2024advanced,
  title={Advanced control techniques for high maneuverability trajectory tracking in autonomous aerial robots},
  author={Alqudsi, Yunes},
  booktitle={2024 1st International Conference on Emerging Technologies for Dependable Internet of Things (ICETI)},
  pages={1--8},
  year={2024},
  organization={IEEE}
}

@article{abro2024synergistic,
  author  = {Abro, G. E. M. and Ali, Z. A. and Masood, R. J.},
  title   = {Synergistic UAV Motion: A Comprehensive Review on Advancing Multi-Agent Coordination},
  journal = {IECE Transactions on Sensing, Communication, and Control},
  volume  = {1},
  number  = {1},
  pages   = {72--88},
  year    = {2024},
  publisher = {IECE}
}

@article{arshid2025toward,
  title={Toward Autonomous UAV Swarm Navigation: A Review of Trajectory Design Paradigms},
  author={Arshid, Kaleem and Krayani, Ali and Marcenaro, Lucio and Gomez, David Martin and Regazzoni, Carlo},
  journal={Sensors (Basel, Switzerland)},
  volume={25},
  number={18},
  pages={5877},
  year={2025}
}

@article{zhang2025cooperative,
  author  = {Z. Zhang and J. Jiang and K. V. Ling and X. Wang and W. A. Zhang},
  title   = {Cooperative Path Planning for Heterogeneous UAV Swarms: A Stackelberg Game Approach},
  journal = {IEEE Transactions on Automation Science and Engineering},
  year    = {2025},
  note    = {Early Access},
  month   = {Jul},
  day     = {4}
}

@article{hafezi2022design,
  author  = {Hafezi, H. and Bakhtiari, A. and Khaki-Sedigh, A.},
  title   = {Design and implementation of a fault-tolerant controller using control allocation techniques in the presence of actuators saturation for a VTOL octorotor},
  journal = {Robotica},
  volume  = {40},
  number  = {9},
  pages   = {3057--3076},
  year    = {2022},
  publisher = {Cambridge University Press}
}

@article{alqudsi2023general,
  author  = {Alqudsi, Y. S. and Kassem, A. H. and El-Bayoumi, G.},
  title   = {A general real-time optimization framework for polynomial-based trajectory planning of autonomous flying robots},
  journal = {Proceedings of the Institution of Mechanical Engineers, Part G: Journal of Aerospace Engineering},
  volume  = {237},
  number  = {1},
  pages   = {29--41},
  year    = {2023},
  publisher = {SAGE Publications}
}

@article{ref20,
  author  = {Lalla-Ruiz, E. and Mes, M.},
  title   = {Mathematical formulations and improvements for the multi-depot open vehicle routing problem},
  journal = {Optimization Letters},
  volume  = {15},
  pages   = {271--286},
  year    = {2021}
}

@article{alqudsi2021robust,
  author  = {Alqudsi, Y. and Kassem, A. and El-Bayoumi, G.},
  title   = {A robust hybrid control for autonomous flying robots in an uncertain and disturbed environment},
  journal = {INCAS Bulletin},
  volume  = {13},
  number  = {2},
  year    = {2021}
}

@mastersthesis{ref21,
  author = {Polychronis, G.},
  title  = {Investigating the Dynamic Multi-Vehicle Routing Problem under Energy Constraints},
  school = {University of Thessaly},
  year   = {2021}
}

@article{gerkey2004formal,
  title={A formal analysis and taxonomy of task allocation in multi-robot systems},
  author={Gerkey, Brian P and Matari{\'c}, Maja J},
  journal={The International journal of robotics research},
  volume={23},
  number={9},
  pages={939--954},
  year={2004},
  publisher={SAGE Publications}
}

@article{song2023survey,
  title={Survey on mission planning of multiple unmanned aerial vehicles},
  author={Song, Jia and Zhao, Kai and Liu, Yang},
  journal={Aerospace},
  volume={10},
  number={3},
  pages={208},
  year={2023},
  publisher={MDPI}
}

@inproceedings{stern2019multi,
  title={Multi-agent pathfinding: Definitions, variants, and benchmarks},
  author={Stern, Roni and Sturtevant, Nathan and Felner, Ariel and Koenig, Sven and Ma, Hang and Walker, Thayne and Li, Jiaoyang and Atzmon, Dor and Cohen, Liron and Kumar, TK and others},
  booktitle={Proceedings of the International Symposium on Combinatorial Search},
  volume={10},
  number={1},
  pages={151--158},
  year={2019}
}

@inproceedings{mellinger2011minimum,
  title={Minimum snap trajectory generation and control for quadrotors},
  author={Mellinger, Daniel and Kumar, Vijay},
  booktitle={2011 IEEE international conference on robotics and automation},
  pages={2520--2525},
  year={2011},
  organization={IEEE}
}

@article{kong2024multi,
  title={Multi-UAV simultaneous target assignment and path planning based on deep reinforcement learning in dynamic multiple obstacles environments},
  author={Kong, Xiaoran and Zhou, Yatong and Li, Zhe and Wang, Shaohai},
  journal={Frontiers in Neurorobotics},
  volume={17},
  pages={1302898},
  year={2024},
  publisher={Frontiers Media SA}
}
